\documentclass[nohyperref]{article}

\usepackage{microtype}
\usepackage{graphicx}
\usepackage{subfigure}
\usepackage{booktabs,multirow} 

\usepackage{hyperref}

\usepackage[accepted]{icml2023}

\usepackage{nicefrac}
\usepackage{amsmath}
\usepackage{amssymb}
\usepackage{mathtools}
\usepackage{amsthm}

\usepackage[capitalize,noabbrev]{cleveref}

\theoremstyle{plain}
\newtheorem{theorem}{Theorem}[section]

\newtheorem{lemma}[theorem]{Lemma}
\newtheorem{corollary}[theorem]{Corollary}
\theoremstyle{definition}
\newtheorem{definition}[theorem]{Definition}
\newtheorem{assumption}[theorem]{Assumption}
\theoremstyle{remark}

\usepackage[textsize=tiny]{todonotes}

\icmltitlerunning{Unveiling the Latent Space Geometry of Push-Forward Generative Models}

\DeclarePairedDelimiterX{\infdivx}[2]{(}{)}{%
  #1\;\delimsize\|\;#2%
}

\DeclarePairedDelimiterX{\infdivy}[2]{(}{)}{%
  #1,#2%
}
\DeclareMathOperator*{\argmax}{arg\,max}
\DeclareMathOperator*{\argmin}{arg\,min}

\usepackage{mathrsfs} 
\usepackage{dsfont}
\usepackage{comment}
\usepackage{enumerate}
\usepackage{tabularx}

\usepackage{natbib}
\bibpunct{(}{)}{;}{a}{,}{,}
\usepackage{hyperref}
\hypersetup{
colorlinks = true,
urlcolor = blue,
linkcolor = blue,
citecolor = blue,
}

\begin{document}

\twocolumn[
    \title{Unveiling the Latent Space Geometry of Push-Forward Generative Models}
    \icmltitle{Unveiling the Latent Space Geometry of Push-Forward Generative Models}


\begin{icmlauthorlist}
    \icmlauthor{Thibaut Issenhuth}{sch,yyy}
    \icmlauthor{Ugo Tanielian}{sch}
    \icmlauthor{Jérémie Mary}{sch}
    \icmlauthor{David Picard}{yyy}
\end{icmlauthorlist}
\icmlaffiliation{yyy}{Ecole des Ponts, UGE, CNRS, France}
\icmlaffiliation{sch}{Criteo AI Lab, Paris, France}
\icmlcorrespondingauthor{Thibaut Issenhuth}{t.issenhuth@criteo.com}
\icmlkeywords{Machine Learning, ICML}
\vskip 0.3in
]
\printAffiliationsAndNotice{}

\begin{abstract}

    Many deep generative models are defined as a push-forward of a Gaussian measure by a continuous generator, such as Generative Adversarial Networks (GANs) or Variational Auto-Encoders (VAEs). This work explores the latent space of such deep generative models. A key issue with these models is their tendency to output samples outside of the support of the target distribution when learning disconnected distributions. We investigate the relationship between the performance of these models and the geometry of their latent space. Building on recent developments in geometric measure theory, we prove a sufficient condition for optimality in the case where the dimension of the latent space is larger than the number of modes. Through experiments on GANs, we demonstrate the validity of our theoretical results and gain new insights into the latent space geometry of these models. Additionally, we propose a truncation method that enforces a simplicial cluster structure in the latent space and improves the performance of GANs. 
\end{abstract}


\section{Introduction}


GANs \citep{GANs} and VAEs \citep{kingmaauto} have shown great capacities to generate photorealistic images \citep{karras2021alias,vahdat2020nvae}. These two models are also helpful for diverse tasks such as image editing \citep{shen2020interpreting, wu2021stylespace} or unsupervised image segmentation \citep{abdal2021labels4free,zoran2021parts}. GANs and VAEs rely on learning a Lipschitz-continuous transformation from a low dimensional Gaussian space. As such, they have been described as \textit{push-forward generative models} \citep{salmonacan}. According to the same taxonomy, score-based models can be defined as \textit{indirect push-forward generative models} since they result from the composition of a large number of transformations and are trained with an auxiliary denoising objective.

The present paper aims at making a step towards a better understanding of push-forward generative models such as GANs. In particular, the goal is to shed light on the latent space of these architectures, and to stress how it impacts the performance of both GANs and VAEs. If empirical studies such as \citet{donahue2019large} have suggested the emergence of simple geometrical structure in the latent space of GANs, there is still a poor theoretical understanding of how generators organize their latent space. 

To better understand the latent space of generative models, the setting of disconnected distributions learning is enlightening. Experimental and theoretical works  \citep{khayatkhoei2018disconnected, tanielian2020learning,salmonacan} have shown a fundamental limitation of push-forward generative models. Since the modeled distribution is connected, some areas of its support are necessarily mapped outside the true data distribution. However, when covering several modes of a disconnected distribution, generators still try to minimize the numbers of samples lying outside the true modes (\textit{e.g.} the purple area on the right of Figure \ref{fig:intro}). In other words, generators aim at minimizing the measure of the existing borders between the modes in the latent space. Considering a Gaussian latent space, finding such minimizers is closely linked to Gaussian isoperimetric inequalities \citep{ledoux1996isoperimetry} where the goal is to derive the partitions that split a Gaussian space with minimal Gaussian-weighted perimeters. Most notably, a recent result \citep{milman2022gaussian} shows that, as long as the number of components $m$ in the partition  and the number of dimensions $d$ of the Gaussian space are such that $m\leq d+1$, the optimal partition is a `simplicial cluster': a Voronoi diagram with equidistant seeds, see left of Figure \ref{fig:intro} for $m=3$ and $d=3$. 


In this paper, we demonstrate the effectiveness of applying simplicial clusters to the latent space of push-forward generative models. We show both experimentally and theoretically that generators with a latent space structured as a simplicial cluster minimize the occurrence of out-of-distribution generated samples. Using the \textit{precision} metric \citep{sajjadi2018assessing, kynkaanniemi2019improved},  
we show that generators with a simplicial cluster latent space achieve optimal precision levels and provide both an upper and a lower bound on their precision. Our experiments reveal that GANs with higher performances tend to organize their latent space as simplicial clusters. More importantly, we illustrate that enforcing this `simplicial structure' with a truncation method can boost GANs' performance. Interestingly, simplicial clusters are highly similar to the `simplex Equiangular Tight Frames' observed in the last-layer features of deep classification networks \citep{papyan2020prevalence}. This study stresses that they also naturally emerge in deep push-forward generative models. Our contributions are the following: 

\begin{figure}
    \centering
    {\includegraphics[width=0.25\linewidth]{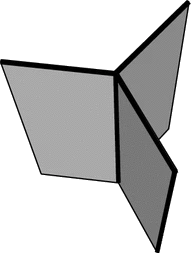}}
    {\includegraphics[width=0.4\linewidth]{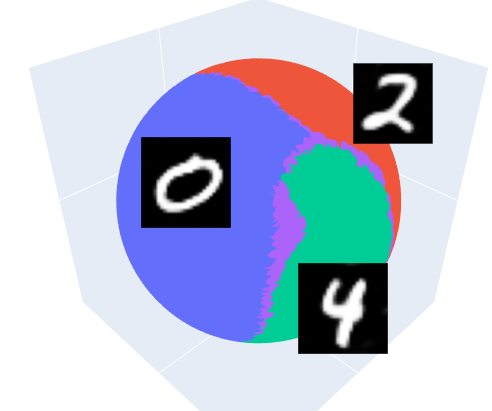}}
        \caption{Illustration of the capability of GANs to discover an optimal geometry of the latent space. On the left, the propeller shape represents a partition of 3D Gaussian space with the smallest Gaussian-weighted perimeter (Figure from \cite{heilman2013solution}). On  the right, we show the 3D Gaussian latent space of a GAN trained on three classes of MNIST. Each area colored in blue, green, or red corresponds to samples in one of the three classes. Using a pre-trained classifier, we highlight in purple the samples with low-confidence, and observe that the partition reached by the GAN (right) is close to optimality (left), as the latent space partition is similar to the intersection of the propeller on a sphere. \label{fig:intro}}
    \end{figure}

\begin{itemize}
    \item We are the first to build on the latest results from Gaussian isoperimetric inequalities by \citet{milman2022gaussian} in the study and understanding of push-forward generative models. 
    \item We present a new theoretical analysis, providing both an upper bound on the precision of push-forward generative models. We demonstrate that generators with a latent space organized as a simplicial cluster have an optimal precision, with lower bounds that decrease in $\sqrt{m\log m}$, where $m$ is the number of modes.
    
    \item Experimentally, we verify that GANs tend to structure their latent space as simplicial clusters' by exploring two properties of the latent space: linear separability and convexity of classes. Also, we analyse the impact of latent space dimension on GANs, and reveal a positive correlation between GANs' performance and latent space geometry.  
    
    \item Finally, we show that enforcing a simplicial structure into GANs' latent space can boost their performance and outperforms other boosting methods. 
\end{itemize}


\section{Related Work}
\subsection{Notation}
\paragraph{Data.}
We consider a target distribution $\mu_\star$ defined on a Euclidean space $\mathds{R}^D$, which may be a high-dimensional space, and equipped with the Euclidean norm $\|\cdot\|$. We use $S_{\mu}$ to represent the support of any distribution $\mu$.


\paragraph{Push-forward generative models.}
We consider the set of $L$-Lipschitz continuous functions, denoted as $\mathscr{G}_L$, from the latent space $\mathds{R}^d$ to the high-dimensional space $\mathds{R}^D$. The primary goal of each generator in this set is to produce realistic samples. The distribution in the latent space, defined on $\mathds{R}^d$, is assumed to be Gaussian and is represented as $\gamma$. For each generator $G \in \mathscr{G}_L$, we associate the push-forward distribution (or image distribution) of $\gamma$ by $G$, and denote it $G \sharp \gamma$, where $\sharp$ denotes the push-forward operator. In the context of generative models, each distribution $G \sharp \gamma$ is now a candidate distribution to represent $\mu_\star$.

The Lipschitzness assumption on $\mathscr{G}_L$ is reasonnable: \citet{virmaux2018lipschitz} have shown the lipschitzness of deep neural networks, and have developed an algorithm that can upper-bound their Lipschitz constant. While deep neural networks can have high Lipschitz constants, it is possible to constrain this in practice by techniques such as clipping the neural network's parameters \citep{arjovsky2017wasserstein}, penalizing the discriminative functions' gradient \citep{gulrajani2017improved, kodali2017convergence, wei2018improving, zhou2019lipschitz}, or penalizing the spectral norms \citep{spectral_normGANs}. Large-scale generators such as SAGAN \citep{zhang2019self} and BigGAN \citep{brock2018large} also make use of spectral normalization for the generator.


\subsection{Generative models and disconnected distributions}
The phenomenon of misspecification in continuous generative models, while primarily studied in the context of GANs, is also relevant to other families such as VAEs or normalizing flows \citep{salmonacan}. This issue has been investigated both experimentally \citep{khayatkhoei2018disconnected} and theoretically  \citep{tanielian2020learning,salmonacan}. The problem stems from a fundamental trade-off: continuous generators can either cover all modes, resulting in out-of-manifold samples, or generate only high-quality samples, neglecting some modes. To address this, various methods have been proposed, such as training disconnected distributions \citep{gurumurthy2017deligan,khayatkhoei2018disconnected} or deriving rejection mechanisms from pre-trained generators \citep{azadi2018discriminator,tanielian2020learning,humayun2022polarity}. 

Empirical studies have provided valuable insights into the structure of the latent space of generative models. For example, \citet{karras2018style} demonstrate that binary attributes are linearly separable in the Gaussian latent space and even more separable in an intermediate latent space. Similarly, \citet{shen2020interpreting} find that face attributes are separated by hyperplanes in the latent space.
\citet{arvanitidis2018latent} and \citet{chen2018metrics} view the latent space of generative models with a Riemannian perspective.

While these findings provide valuable insights into the latent space structure of generative models, they may not be sufficient for a comprehensive understanding of the latent space geometry. For instance, 
\citet{tanielian2020learning} stress the relevance of this problem by showing that the precision of GANs can converge to $0$ when the number of modes or the distance between them increases. In this paper, we take a step towards a deeper understanding of the behavior of push-forward generative models and reveal an optimal latent space configuration when the number of modes $m$ and the dimension of the latent space $d$ are such that $m \leq d+1$.

\subsection{Evaluating generative models}
When learning disconnected manifolds, \citet{sajjadi2018assessing} illustrated the need for measures that simultaneously evaluate both the quality (Precision), and the diversity (Recall) of the generated samples. However, \citet{kynkaanniemi2019improved} pointed out an important limitation of the PR metric: it cannot accurately interpret situations when large numbers of samples are packed together. They propose an Improved PR metric based on the non-parametric estimation of manifolds to correct this. 

\paragraph{Improved PR metric.} 
Informally, for a generator $G$, precision ($\alpha_G$) quantifies the proportion of generated samples that can be approximated with true samples, while recall ($\beta_G$) measures the proportion of true samples that can be approximated with generated ones. Applying this to GANs, using the target distribution $\mu^\star$ and modeled distribution $G \sharp \gamma$, the Improved PR metric was shown, by \citet[][Theorem 1]{tanielian2020learning}, to be asymptotically equivalent to:
\begin{equation*}
    \alpha^n_G \underset{n \to \infty}{\rightarrow} \alpha_G = G \sharp \gamma \big(S_{\mu^\star} \big) \ \text{and} \ \beta^n_G \underset{n \to \infty}{\rightarrow} \beta_G = \mu^\star \big(S_{G \sharp \gamma}\big), 
\end{equation*}
where $S_{\mu^\star}$ denotes the support of $\mu^\star$ and $n$ is the number of samples. However, \citet{naeem2020reliable} have shown that the Improved PR metric \citep{kynkaanniemi2019improved} is sensitive to outlier samples of both the target and the generated distribution. To correct this and fix the overestimation of the manifold around real outliers, \citet{naeem2020reliable} propose the Density/Coverage metric. 



\paragraph{Density/Coverage.} 
Instead of counting how many fake samples belong to a real sample neighborhood, density counts how many real sample neighborhoods contain a generated sample. On the other hand, coverage counts the number of real sample neighborhoods that contain at least one fake sample. 

In the next analysis both theoretical and experimental, we use both notions of precision and density defined above. 

\section{Simplicial Structure in Push-Forward Generative Models}
The goal is to gain a deeper understanding of the latent space of push-forward generative models and identify which ones possess the highest precision under certain conditions. As previously mentioned, push-forward generative models map a unimodal Gaussian distribution $\gamma$ through a Lipschitz-continuous function, represented by a generator $G$. As a result, the modeled generative distribution $G \sharp \gamma$ necessarily has a connected support.

In cases where the target distribution $\mu^\star$ contains disconnected manifolds, generators have to generate fake data points that fall outside of the true manifold. This prompts the question: given that a generator samples data points from each of the distinct modes, what is the maximum precision that it can achieve? To begin with, let's consider a target distribution $\mu^\star$ composed of $m$ disconnected modes.

\begin{assumption}[Disconnected manifolds]
\label{ass:disconnect} 
    The target distribution $\mu^\star$ consists of $m$ disconnected spheres $S_i, i \in [1,m]$ of equal measure (with centers $X_i$ and radius $r_i$). Additionally, the spheres satisfy the two following properties: 
    \begin{itemize}
        \item \textbf{Small individual radius:} each radius $r_i$ satisfies 
            \begin{equation}\label{radius}
                r_i < \min_{j}  \frac{\|X_i - X_j\|}{2}.        
            \end{equation}
        \item Each distance $\|X_i - X_j \|$ satisfies:
    \begin{equation}\label{eq:nn_middle}
        \min_{k \in [1,m], k \neq i,j} \ \| (X_i+X_j)/2 - X_k \| > \frac{\|X_i - X_j \|}{2}.
    \end{equation}
    \end{itemize}
\end{assumption}
We believe that the assumption of disconnectedness is a reasonable one, particularly for multi-class datasets such as MNIST \cite{lecun98gradientbasedlearning}, CIFAR10 \cite{krizhevsky2009learning}, or STL10 \cite{coates2011analysis}. To validate this property, we run a pre-trained CLIP \citep{radford2021learning} on the dataset, identify a certain number of clusters using a K-means algorithm, and further test the disconnectedness of these modes by training a linear classifier. The accuracy on these datasets is 98.1\% on MNIST, 93.9\% on CIFAR10, and 92.7\% on STL10. 
The second point in \eqref{eq:nn_middle} has a direct impact on the location of the data points $X_1, \hdots, X_m$. Specifically, it implies that each cell in the Voronoi diagram with seeds $X_1, \hdots, X_m$ shares a side with all the other cells. In other words, the dual graph of this Voronoi diagram is complete. This assumption, which is further discussed with specific examples in Figure \ref{fig:three_points_aligned}, can be justified by the concentration of distances in high-dimensional spaces: all the modes are roughly at equal distance \citep{beyer1999nearest,aggarwal2001surprising}. Furthermore, a recent work by \citet{papyan2020prevalence} has shown that embeddings of deep neural networks trained for classification tend to collapse around means that are equidistant and maximally equiangular to one another. By using these embedded representations to measure distance, the target distribution would thus easily satisfy Assumption \ref{ass:disconnect}. Projected GANs \citep{sauer2021projected} is really close to this idea as the authors show the effectiveness of leveraging a pre-trained classifier when training GANs: instead of directly discriminating images, the discriminator is trained on features extracted from the classifier.
Throughout the rest of the paper, we define the set of \textit{well-balanced} generators as those mapping an equal number of data points to each mode of the data distribution:
\begin{definition}\label{def:well-balanced}
A generator $G$ is well-balanced if for all spheres, we have $G \sharp \gamma (S_1) = \hdots = G \sharp \gamma (S_m)$.
\end{definition}
Considering well-balanced generators is reasonable as many empirical improvements such as WGAN-GP \citep{gulrajani2017improved} or BigBiGAN \citep{donahue2019large} have significantly reduced mode collapse. GANs generate diverse output distributions on datasets such as CIFAR10, CIFAR100, and ImageNet. To validate the use of well-balanced generators, we conducted a small experiment and evaluated the proportion of each class generated by GANs on MNIST and CIFAR10. On MNIST, the minimum proportion of a class is $9.2$ and the maximum $10.9$, while on CIFAR10 it is $8.3$ and $11.9$ (in \%). The variance-to-mean ratio is equal to $0.03$ for MNIST and $0.22$ for CIFAR10.

\subsection{Precision and the associated partition}
Now that we have defined the prerequisites for both the data and the model, we propose to establish a connection between the latent space partition and the precision of a generator. We create a link between the set of generators from $\mathds{R}^d$ to $\mathds{R}^D$ and the set of partitions in the latent space. Specifically, for each given partition in $\mathds{R}^d$, there exists a set of associated generators defined as follows:
\begin{definition}\label{def:association}
For a given partition $\mathcal{A}={A_1, \hdots, A_m }$ on $\mathds{R}^d$, we say that $G$ is associated to $\mathcal{A}$ if: for all $i \in [1,m], \ \text{for all} \ z \in A_i, \ i = \underset{j \in [1,m]}{\argmin} \ \|G(z)-X_j\|.$
\end{definition}
Each given generator $G$ is associated with a unique partition $\mathcal{A}$ in $\mathds{R}^d$. The geometry of the associated partition $\mathcal{A}$ plays a key role in explaining the behavior and performance of the generator $G$. We are interested in maximizing the precision of generative models. Points in the intersection of two cells $A_i \cap A_j, (i,j) \in [1,m]^2$ are equidistant from $X_i$ and $X_j$ and thus do not belong to any of these modes (since bot $r_i$ and $r_j < \|X_i - X_j\|/2$ according to Assumption \ref{ass:disconnect}). Additionally, due to the generator's Lipschitzness, there is a small neighborhood around the boundary such that any points in this neighborhood are mapped out of the target manifold. This region in the latent space thus reduces the precision. For a given $\varepsilon > 0$, we now define the epsilon-boundary of the partition $\mathcal{A}$ as follows.
\begin{definition}\label{def:epsilon_boundary}
    For a given partition $\mathscr{A}=\{A_1, \hdots, A_m\}$ of $\mathds{R}^d$ and a given $\varepsilon \in \mathds{R}^\star_{+}$, we denote $\partial^\varepsilon \mathcal{A}$ the $\varepsilon$-boundary of $\mathcal{A}$, defined as follows.
    \begin{equation*}
        \partial^\varepsilon \mathcal{A} = \bigcup_{i=1}^m \big( \cup_{j \neq i} A_j\big)^\varepsilon \backslash \big(\cup_{j \neq i} A_j \big),
    \end{equation*}
\end{definition}
where $A^\varepsilon$ corresponds to the $\varepsilon$-extension of set $A$. The following lemma makes the connection between the precision of a generator $\alpha_G$ and its associated partition $\mathcal{A}$.
\begin{lemma}\label{lem:11}
    Assume that Assumption \ref{ass:disconnect} is satisfied and $\mathscr{A}$ be a partition in $\mathds{R}^d$. Then, any generator $G \in \mathcal{G}_{L}$ associated with $\mathcal{A}$ verifies:
    \begin{equation}\label{eq:precision_boundary}
        \alpha_G \leqslant 1 - \gamma (\partial^{\varepsilon_{\text{min}}} \mathcal{A}).
    \end{equation}
    where $\varepsilon_{\text{min}} = \min_{i,j} \|X_i-X_j\|/L$.
\end{lemma}
Interestingly, this result holds independently of the partition $\mathcal{A}$. It highlights that the geometry of the partition gives an upper-bound on the precision of the generator. Consequently, to properly determine this bound on the precision levels of generative models, one might be interested in determining the measure of this epsilon-boundary $\partial^\varepsilon \mathcal{A}$. By using the result from Lemma \ref{lem:11}, we can derive an upper-bound on the precision that depends on $D, L$ and $m$:
\begin{corollary}\label{cor:cor36}
    Assume that Assumption \ref{ass:disconnect} is satisfied, $m\leqslant d+1$. Then, there exists $L$ with $L \geqslant D \sqrt{\log(m)}$, such that for any well-balanced generator $G \in \mathscr{G}_L$:
    \begin{equation}\label{eq:precision_finite}
        \alpha_G 
        \leqslant 1 - \varepsilon_{\text{min}} \sqrt{\log m} \ e^{-3/2}
    \end{equation}
\end{corollary}

where $\varepsilon_{\text{min}} = \min_{i,j} \|X_i-X_j\|/L$. In particular, the result in \eqref{eq:precision_finite} gives an interesting insight when training GANs on a finite number of modes. \citet[][Theorem 3]{tanielian2020learning} showed a similar result but for the asymptotic case when the number of modes increases:
\begin{equation}\label{eq:precision_asymptotic}
    \alpha_G \overset{m \rightarrow \infty}{\leqslant} e^{-\frac{1}{8}\varepsilon_{\text{min}}^2}e^{-\varepsilon_{\text{min}} \sqrt{\log(m)/2}}.
\end{equation}

\subsection{Optimality for push-forward generative models}
To exhibit generative models with optimal precision levels, one must look at partitions with the smallest epsilon-boundary measures $\gamma(\partial^\varepsilon \mathcal{A})$. We argue that this is tightly connected to the theoretical field of Gaussian isoperimetric inequalities. Isoperimetric inequalities link the measure of sets with their perimeters. More specifically, these inequalities highlight minimizers of the perimeter for a fixed measure, \textit{e.g.} the sphere in an euclidean space with a given Lebesgue measure. In the Gaussian space, \citet{borell1975brunn} and \citet{sudakov1978extremal} show that in a finite-dimensional case, among all sets of a given measure, half-spaces have a minimum Gaussian perimeter. More formally, for any Borel set $A$ in $\mathds{R}^d$ and a half-space $H$, if we have $\gamma({A}) \geqslant \gamma(H)$, then $\gamma({A}^\varepsilon) \geqslant \gamma(H^\varepsilon)$ for any $\varepsilon > 0$, where ${A}^\varepsilon$ denotes the $\varepsilon$-extension of $A$. 

The Gaussian multi-bubble conjecture was formulated when looking for a way to partition the Gaussian space in $m$ parts, with the least-weighted boundary. It was recently proved by \citet{milman2022gaussian} who showed that the best way to split a Gaussian space $\mathds{R}^d$ in $m$ clusters of equal measure, with $2 \leqslant m \leqslant d+1$, is by using ‘simplicial clusters' obtained as the Voronoi cells of $m$ equidistant points in $\mathds{R}^d$. Convex geometry theory tells us that each cell is a convex cone, whose borders are hyperplanes going through the origin of $\mathds{R}^d$. We note \textbf{$\mathcal{A}^\star$} any partition corresponding to this optimal configuration, see Figure \ref{fig:intro} for $m=3$. 

In the following theorem, we apply this result to the understanding of GANs. We make the connection between optimal generators (when $m \leqslant d+1$) in levels of precision and the partition $\mathcal{A}^\star$ derived in \citet{milman2022gaussian}. 


\begin{theorem}[\textbf{Optimality of generators with simplicial cluster latent space.}]\label{th:optimal_generators}
     Assume that Assumption \ref{ass:disconnect} is satisfied and $m \leqslant d+1$. For any $\delta>0$, there exists $C$ large enough (independent of $\delta$) and $L \geqslant D \sqrt{m} \sqrt{\pi \log(Cm)}$, and a well-balanced generator $G^\star \in \mathscr{G}_L$ associated with $\mathcal{A}^\star$ such that for any other well-balanced generator $G \in \mathscr{G}_L$, we have:
     \begin{equation}\label{eq:optimality_GANs}
         \alpha_{G^\star}  \geqslant \alpha_G - \delta
    \end{equation}
    Moreover, if $m \leqslant d$, noting $\varepsilon_{\text{max}} = \ \max_{i,j} \|X_i - X_j \|/L$:
    \begin{equation}\label{eq:lower_bound_GANs}
        \alpha_{G^\star} \geqslant 1 - \varepsilon_{\text{max}} \sqrt{m \log(Cm)}, 
    \end{equation}
\end{theorem}


Theorem \ref{th:optimal_generators} shows that when $L$ is large enough, the bound in \eqref{eq:precision_finite} is almost tight, and thus that the given generator based on the simplicial partition $\mathcal{A}^\star$ is almost optimal. However, it is not clear whether those are the only generators with optimal precision. The proof is delayed in Appendix \ref{appendix:proofs}.

\begin{figure}[h]
    \centering
    {{   
        \includegraphics[width=0.33\linewidth]{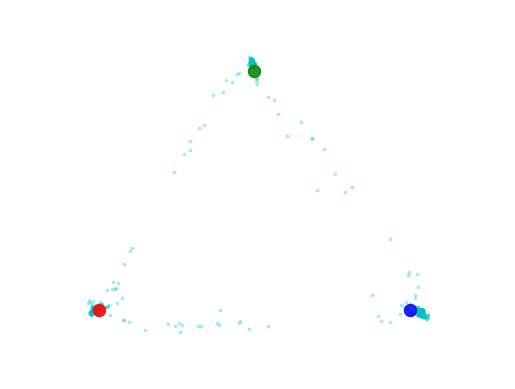}
    }}
    \hfill
    {{   
        \includegraphics[width=0.24\linewidth]{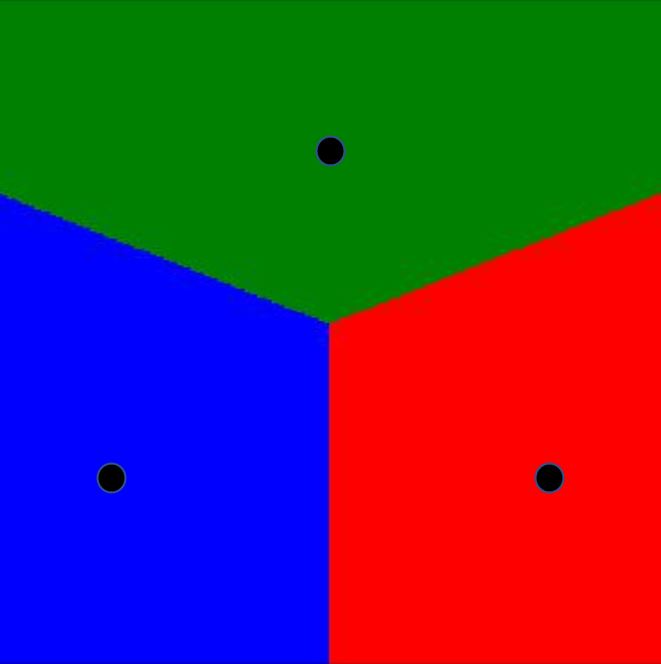}
    }}
    \hfill
    {{   
        \includegraphics[width=0.34\linewidth]{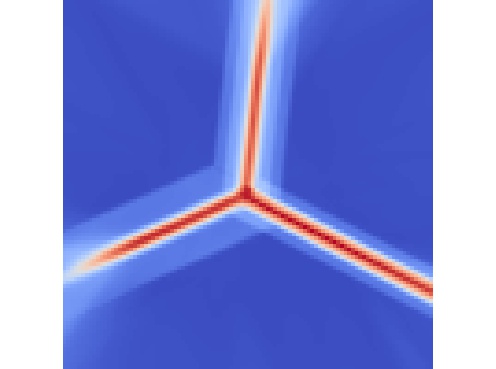}
    }}
    \\
    {{   
        \includegraphics[width=0.33\linewidth]{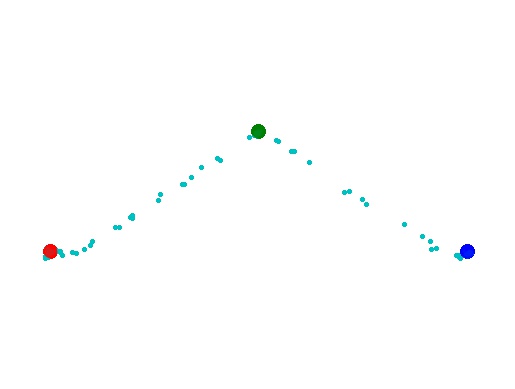}
    }}
    \hspace{0.0192cm}
    {{   
        \includegraphics[width=0.24\linewidth]{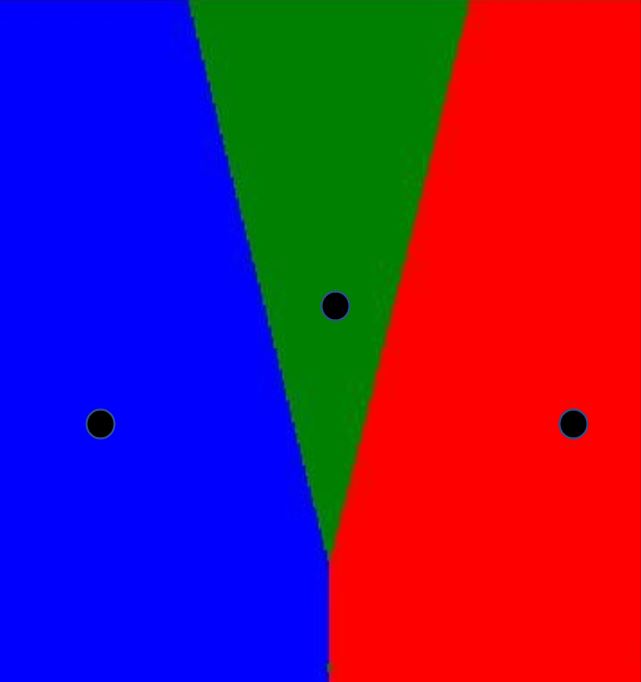}
    }}
    \hspace{0.005cm}
    {{   
        \includegraphics[width=0.34\linewidth]{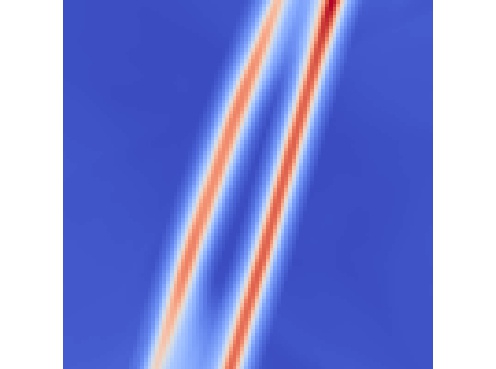}
    }}
    \caption{Illustration of the impact of the geometry of data modes on the latent space of GANs. 
    The left column shows the modes $(X_1, X_2, X_3)$ from the target distribution and the generated points (small blue dots). In the middle, we plot the Voronoi diagram generated from $(X_1, X_2, X_3)$. On the right column, we show the boundaries in the GANs latent space with heatmaps of the norm of the gradient of the generator. In the first row, when the data satisfies Assumption \ref{ass:disconnect}, GANs achieve the optimal configuration. However, when the data modes do not satisfy this assumption, as seen in the second row, this is no longer the case. 
    \label{fig:three_points_aligned}}
\end{figure}

 
\paragraph{What if Assumption \ref{ass:disconnect} is not verified?}
This assumption is needed for the definition of a well-balanced generator associated with $\mathcal{A}^\star$ as in Theorem \ref{th:optimal_generators}. As shown in Figure \ref{fig:three_points_aligned}, the latent space configuration obtained by the GANs for 3 almost equidistant points (1st row) and 3 almost aligned data points (2nd row). We see that in the later case, the Voronoi partition of the target data points does not verify Assumption \ref{ass:disconnect}, and the optimal latent structure is not known. We observe in this specific case that it is made of two parallel hyperplanes, much different from $\mathcal{A}^\star$ defined by \citet{milman2022gaussian} (1st row). 

\begin{figure}[h]
    \centering
    {{   
        \includegraphics[width=0.31\linewidth]{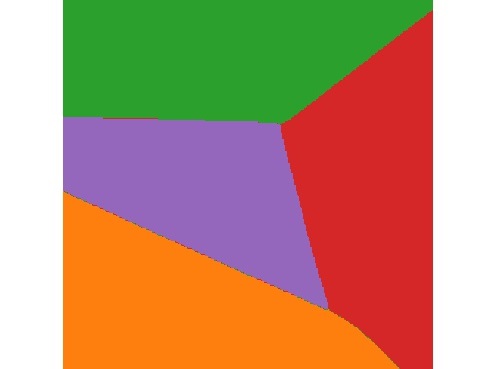}
    }}
    {{   
        \includegraphics[width=0.31\linewidth]{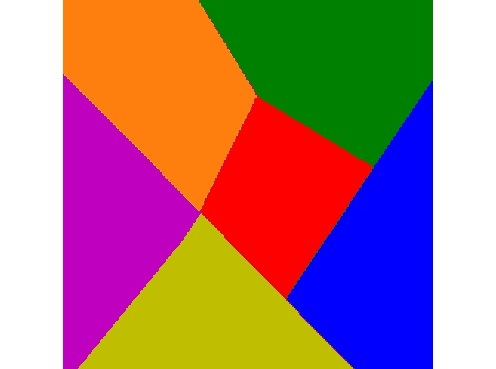}
    }}
    {{   
        \includegraphics[width=0.31\linewidth]{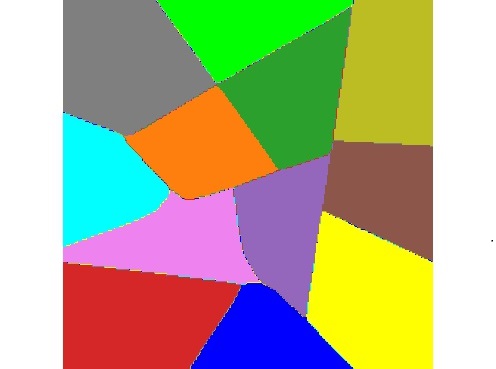}
    }}
    \caption{Extension of the multi-bubble conjecture when $m>d+1$. We depict the partition of the $\mathds{R}^2$ latent space of a GAN that maps to $m$ equidistant points in $\mathds{R}^m$, with $m=4, 6, 12$. Each colored cell maps to a distinct data point in $\mathds{R}^m$. 
    \label{fig:synthetic_gans}}
\end{figure}

\paragraph{What if the dimension $m>d+1$?} 
The position of the different spheres could be such that Assumption \ref{ass:disconnect} is no longer valid. Second, since the result from \citet{milman2022gaussian} does not hold, the optimal partition of the Gaussian space in $m$ equal cells is unknown. In this generalized context, GANs could hint at the optimal partition geometry. Figure \ref{fig:synthetic_gans} stresses examples when training GANs from $\mathds{R}^2$ to $\mathds{R}^m$ with $m$ equidistant modes. This gives some insights on how to divide the Gaussian space into $m$ equitable areas with least Gaussian-weighted perimeter.


\vspace{-0.4cm}
\paragraph{What if the modes do not have equal measure?} 
The fact that each mode has equal measure in the target distribution might not be verified for unbalanced datasets. 
First, the optimality of simplicial clusters holds because the multi-bubble theorem is still valid. However, the lower-bound (Equation \ref{eq:lower_bound_GANs}) does not hold. Additionally, the upper-bound from Corollary \ref{cor:cor36} can be relaxed.  
Consider $w_1, \hdots, w_m \in \mathds{R}^m$ the weights of the different modes, and $w_{\min} = \underset{i}{\min} \ w_i$, the upper-bound becomes:
\begin{equation*}
    \alpha_G 
    \leqslant 1 - m \varepsilon_{\min} w_{\min} \sqrt{\log(1/w_{\min})} \ e^{-3/2}.
\end{equation*}
We observe that this upper-bound might not be tight anymore since it depends on the minimum of the weights  $w_{\min}$.

 \subsection{Improving generative models}\label{sec:improvingGANs}
Our proposed theoretical analysis offers valuable insights into the optimal structure of the latent space for push-forward generative models. We demonstrate that by leveraging this structure, it is possible to design GANs with improved performance.  To achieve this, we enforce a simplicial cluster structure in the latent space of GANs during training using a novel rejection sampling procedure called \textit{simplicial cluster truncation} that can be combined with a mutual-information loss. Note that modifying the latent space distribution of other generative models, such as VAEs or score-based models, is a more complex task.

\paragraph{Simplicial cluster truncation.}
Let us denote a simplicial cluster \citep{milman2022gaussian} as ${(u_1, \hdots , u_m) \mid u_i \in \mathds{R}^d}$. The rejection sampling procedure, based on Theorem \ref{th:optimal_generators}, involves sampling a latent vector $z$ from $\gamma$ and accepting it if $\underset{i \in [1,\hdots,m]}{\max} (z \cdot u_i) > \tau$, where both $\tau$ and $m$ are considered as hyper-parameters. This defines a new latent space distribution where the density is high near the unit vectors $u_i, i \in [1,m]$. As a result, the boundaries of the simplicial cluster, which are points with high distances to the centers of Voronoi cells, are rejected. The threshold parameter $\tau$ determines the $\varepsilon$ value. With this method, the boundaries between different modes are never sampled, leading to a disconnected latent space. This approach can improve the learning of disconnected manifolds by injecting disconnectedness into the modeled generative distribution. Additionally, the use of a geometrical structure that is particularly well suited to separate several modes \citep{papyan2020prevalence} enhances the performance.


\paragraph{Mutual-information loss.} 
The rejection sampling procedure might not be sufficient for the generator to properly use the different clusters of its latent space. To encourage the simplicial cluster structure, we also optimize the mutual information between generated samples and the corresponding cluster \citep{khayatkhoei2018disconnected}. The loss is applied at the beginning of the training and is then dropped.

\section{Experiments} 
In the following experiments, we validate our theoretical analysis and derive insights for GANs trained on toy and image datasets. We verify: 1) that the latent space geometry of GANs has similar properties than simplicial clusters; 2) that increasing the latent space dimension ($d+1>m$) can help improve GANs, as highlighted in the theoretical section; 3) that GANs' performance is correlated with their latent space geometry; 4) that the proposed \textit{simplicial cluster} truncation method is effective and boost GANs' performance. 

In the following experiments, we train WGANs with gradient penalty  \citep{arjovsky2017wasserstein, gulrajani2017improved}. For mixture of Gaussians, generator and discriminator are MLP networks. 
For MNIST, the generator and discriminator are standard convolutional architectures. On CIFAR-10, CIFAR-100 , and STL-10, we use either a Resnet-based \citep{he2016deep} convolutional architecture with self-modulation in the generator \citep{chen2018self}, either the transformer-based architecture from \citet{jiang2021transgan}. To evaluate the performance of GANs, we use both the precision \citep{kynkaanniemi2019improved}, the FID \citep{heusel2017GANs}, and the density/coverage \citep{naeem2020reliable}. We use a dataset-specific classifier to extract image features on MNIST, and InceptionNet pre-trained on ImageNet for CIFAR-10, CIFAR-100 and STL-10. Implementation details are given in Appendix \ref{appendix:experiments} and code is provided in Supplementary Material.

\begin{table*}[h]
\centering
\begin{tabular}{lccccc}
Dataset & Architecture & Latent dim & Precision ($\uparrow$) & LogReg Acc.  ($\uparrow$) & Convex Acc.  ($\uparrow$)\\\toprule
100 Gauss. & MLP & 100 & 75.5 & 78.5 & 87.2  \\
MNIST & CNN & 64 & 93.2 & 90.4 & 98.7 \\
CIFAR-10 & ResNet & 64 & 66.8 & 65.3 & 75.2 \\ 
CIFAR-10 & Transformer & 256 & 72.8  & 70.7 & 84.3  \\ 
CIFAR-100 & ResNet & 64 & 64.3 & 30.5 & 42.1 \\  
CIFAR-100 & Transformer & 64 & 64.2 & 26.5 & 39.2 \\  
\end{tabular}
\caption{Validation of linear separability (LogReg Acc.) and convexity (Convex Acc.) in GAN latent spaces. The results align with the predictions of Corollary \ref{cor:cor36}, where a linearly separable and convex structure of the latent space indicates a high precision. The architecture \textit{Transformer} refers to the TransGAN model from \citet{jiang2021transgan}. The supervised classifiers used as oracles haves test-accuracies of 80.2\% on CIFAR-10 and 61.8\% on CIFAR-100. \label{tab:convexity}}
\end{table*}




\subsection{Linear separability and convexity}
According to \citet{milman2022gaussian}, the optimal configuration in the latent space is obtained as the Voronoi cells of $m$ equidistant points in $\mathds{R}^d$, if $m\leq d+1$. This means that if GANs reach this optimal configuration, each of the cells must be \textit{convex polytopes} and have the following properties: 1) the boundaries of a cell are flat; 2) each cell is convex.
To investigate this, we use a labeled dataset and assess whether a simple linear model (\textit{e.g.}, multinomial logistic regression) can map latents to labels. If the cells in the latent space are bounded by hyperplanes, then, using the hyperplane separation theorem, the linear model is expected to be a good predictor of a generated sample's label.

\begin{figure}
    \centering
    {   
        \includegraphics[width=0.95\linewidth]{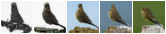}
    }
    \vspace{-0.4cm} \\ 
    {   
        \includegraphics[width=0.95\linewidth]{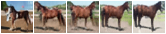} 
    }
    \vspace{-0.4cm} \\ 
    {   
        \includegraphics[width=0.95\linewidth]{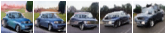}
    }
    \vspace{-0.3cm} \\ 
    \caption{\label{fig:convexity} Visualization of the convexity of classes in the latent space of GANs trained on CIFAR-10. The plot shows that latent linear interpolations within a class preserve the class label. }
\end{figure}

We use a standard multi-class labeled dataset. $G_\theta$ is a pre-trained generator and $C_\phi$ is a pre-trained classifier considered as an oracle. Using $G_\theta$ and $C_\phi$, we construct a dataset of latent vectors $z \in \mathds{R}^d$ and their associated labels $y = C_\phi(G_\theta(z))$. On CIFAR-10/100, similarly to \citet{razavi2019generating}, only data points with a confidence threshold of 0.7 or higher are accepted. This dataset is later split into 100k training points and 10k test points. We use multinomial logistic regression to learn the mapping from latent vectors $z$ to their labels $y$. We can see in Table \ref{tab:convexity}, that the \textit{LogReg Accuracy} reaches high levels: 90\% on MNIST and 70\% on CIFAR-10. For the Convex accuracy, we draw two random latent vectors $z_0$ and $z_1$ that belong to the same class, and check whether linear interpolations in the latent space also belong to the same class, that is $C_\phi(G_\theta(z_0)) = C_\phi(G_\theta(z_0)) = \lambda \times C_\phi(G_\theta(z_0)) + (1-\lambda) \times C_\phi(G_\theta(z_1))$ for $\lambda \in [0,1]$. Interestingly, we see in Table \ref{tab:convexity} a correlation between the Logreg and Convex accuracy and the precision metric: the more the latent space behaves like a simplicial cluster, the higher the precision. For a qualitative evaluation, we show this phenomenon in Figure \ref{fig:convexity} and stress that linear interpolations conserve the image class.

\subsection{Impact of the latent space dimension}\label{sec:impact_latent_dim}
\begin{figure}[h]
    \centering
    {   
        \includegraphics[width=0.475\linewidth]{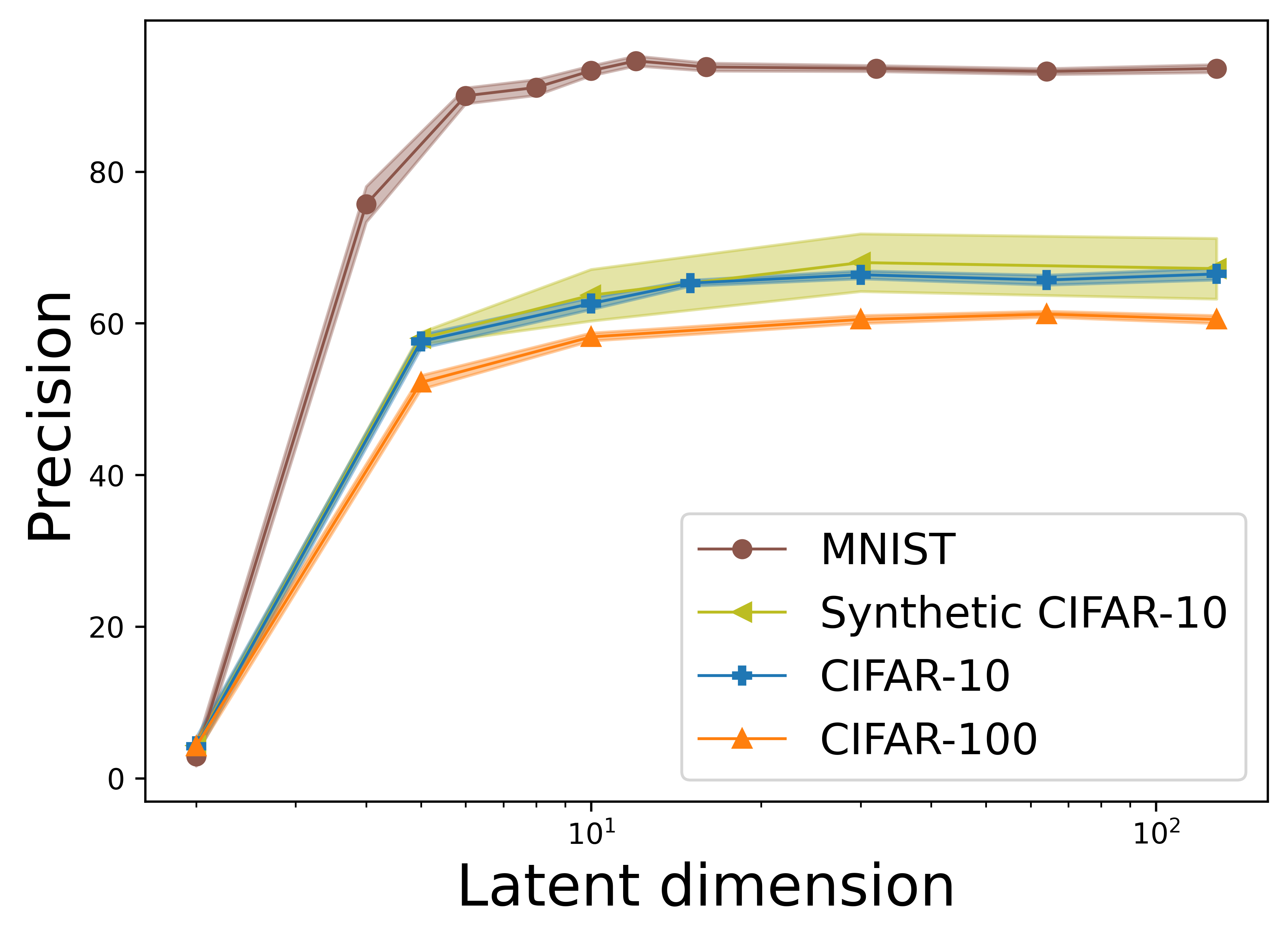}
    }
    {   
    \includegraphics[width=0.475\linewidth]{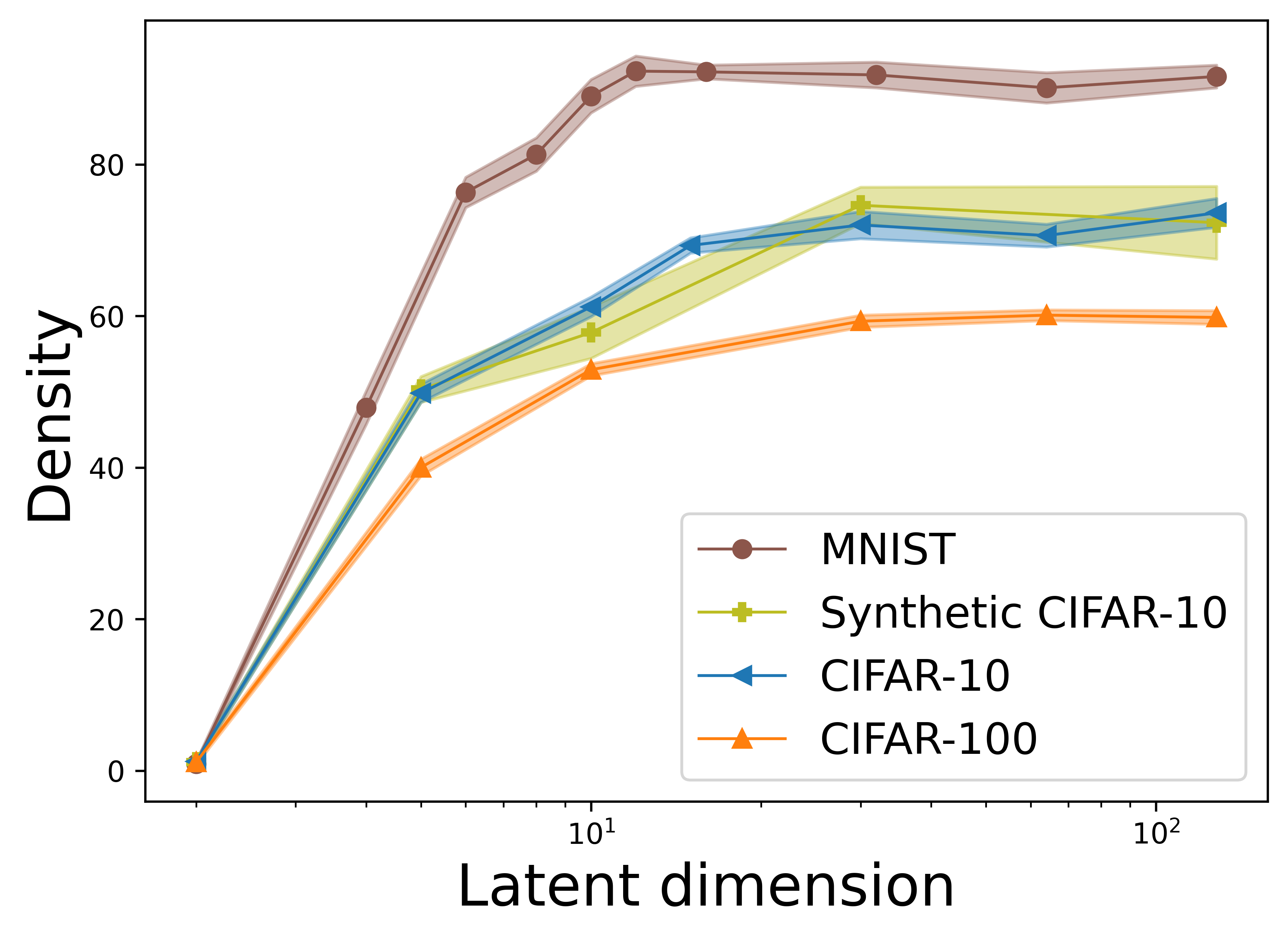}
    }\hfill
    \caption{Performance of GANs with regard to the number of modes and latent space dimensions. As the number of modes and latent space dimension increases, we observe an improvement in Precision (left) and Density (right), with a saturation point beyond a certain threshold.  \label{fig:phase_transition}}
\end{figure}

To evaluate the impact of the latent space dimension, we train GANs with latent space dimension ranging from 2 to 128 on several datasets. 
In Figure \ref{fig:phase_transition}, we exhibit two phases in the performance of GANs when changing the number of latent dimensions. For a fixed architecture, and a given dataset, we observe the existence of an optimal latent space dimension $d^\star$. When $d<d^\star$ the precision or density of the model falls significantly. Interestingly, when $d>d^\star$, the precision becomes constant: overparameterizing the model does not bring a significant improvement. As expected, we observe in Figure \ref{fig:phase_transition} that the maximum precision/density depends on the complexity of the dataset and its number of modes: the more complex the dataset, the lower the precision. This is also coherent with our theoretical results from both Corollary \ref{cor:cor36} and Theorem \ref{th:optimal_generators}.

An interesting problem was also brought to the fore by \citet{roth2017stabilizing}. When training GANs two different issues can arise: 1) \textit{dimensional misspecification} where the true and modeled distributions do not have density functions w.r.t. the same base measure, and 2) \textit{density misspecification}, where GANs try to fit a disconnected manifold with a unimodal distribution. To isolate the density misspecification studied in this paper, we train a conditional GAN with a low-dimensional latent space $\mathds{R}^d$ (\textit{e.g.} $\mathds{R}^5$ in our setting), so that the dimension of the generated manifold is at most 5. We later collect a dataset of synthetic generated samples \textit{Synthetic CIFAR-10}, and train unconditional GANs with varying latent space dimensions. Figure \ref{fig:phase_transition} shows that GANs converge to the same limits for Precision and Density on Synthetic CIFAR-10 and CIFAR-10. This shows that the performance is more impacted by the \textit{density misspecification} (trying to fit a disconnected target distribution with a connected one) rather than the dimensional misspecification. 

\begin{figure*}[t!]
    \centering
    {   \includegraphics[width=0.235\linewidth]{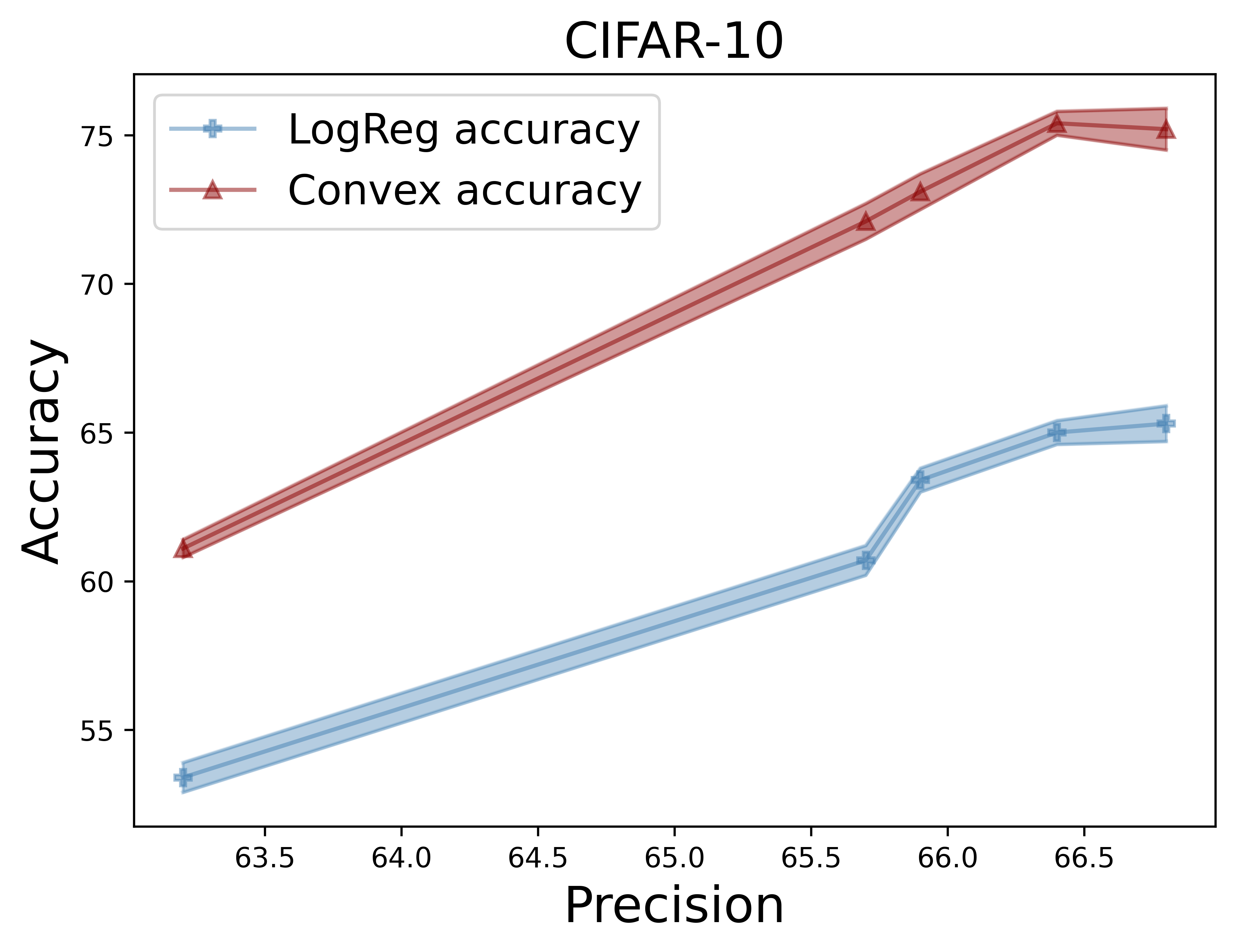}
    }
    {   \includegraphics[width=0.235\linewidth]{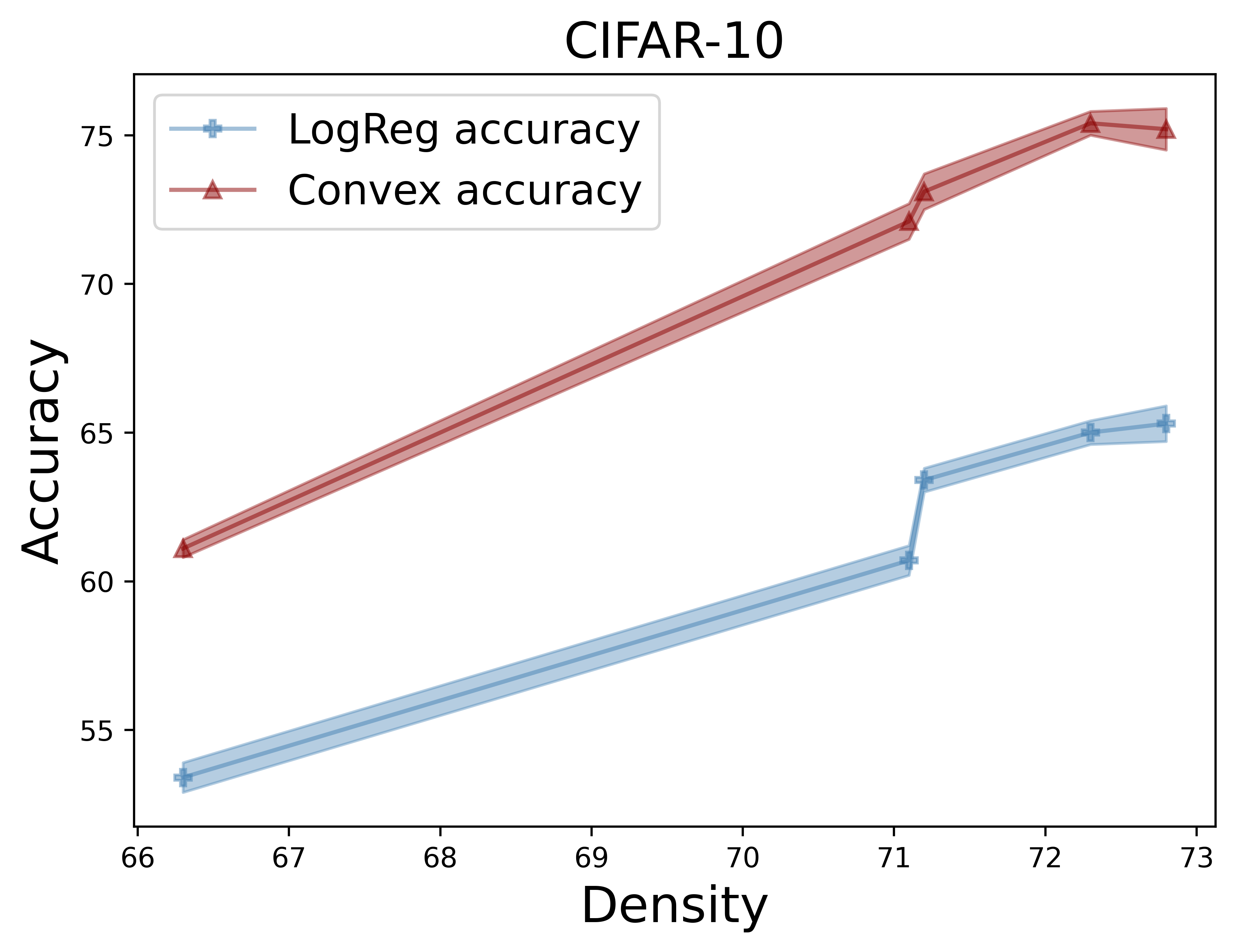}
    }
    {   \includegraphics[width=0.235\linewidth]{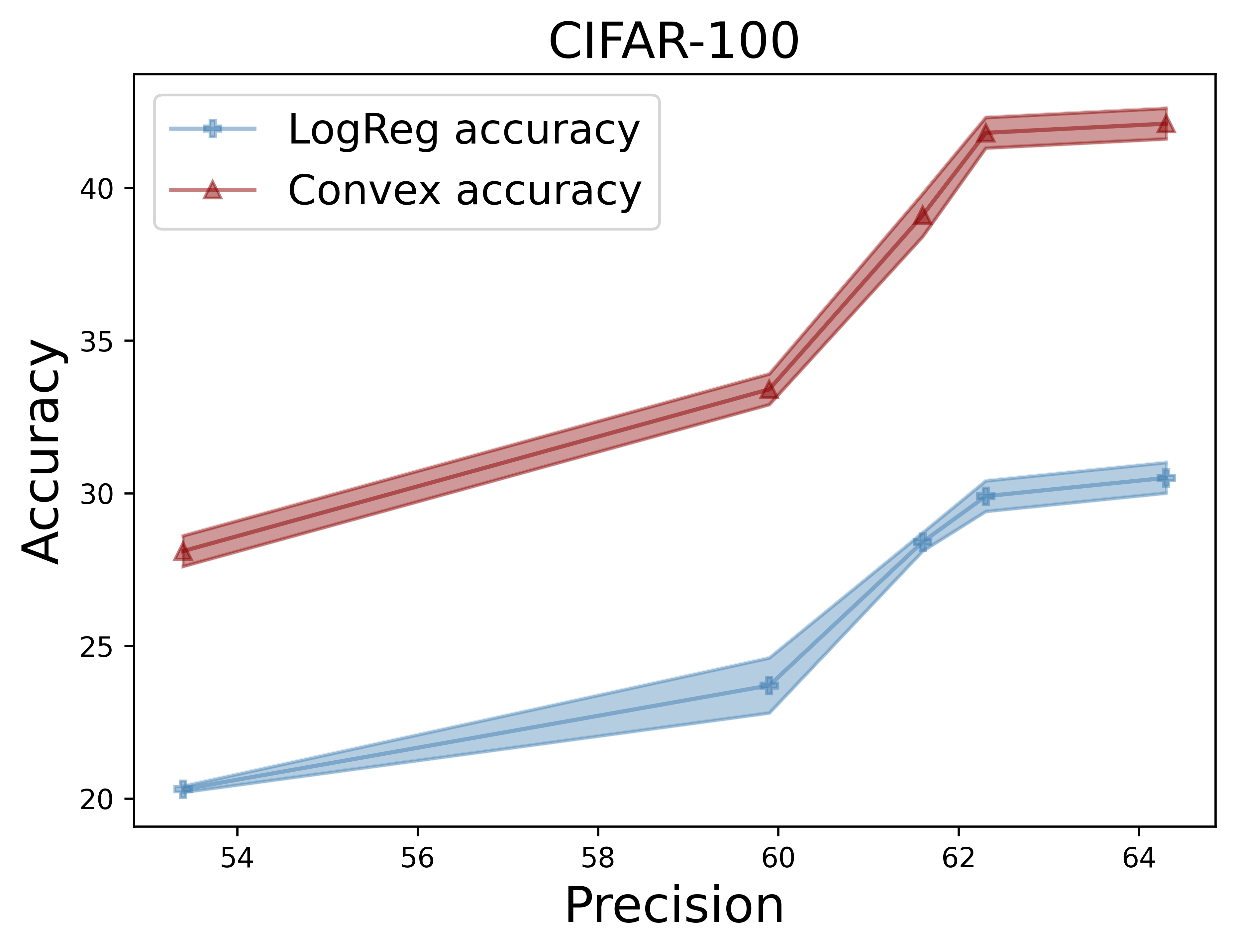}
    }
    {   \includegraphics[width=0.235\linewidth]{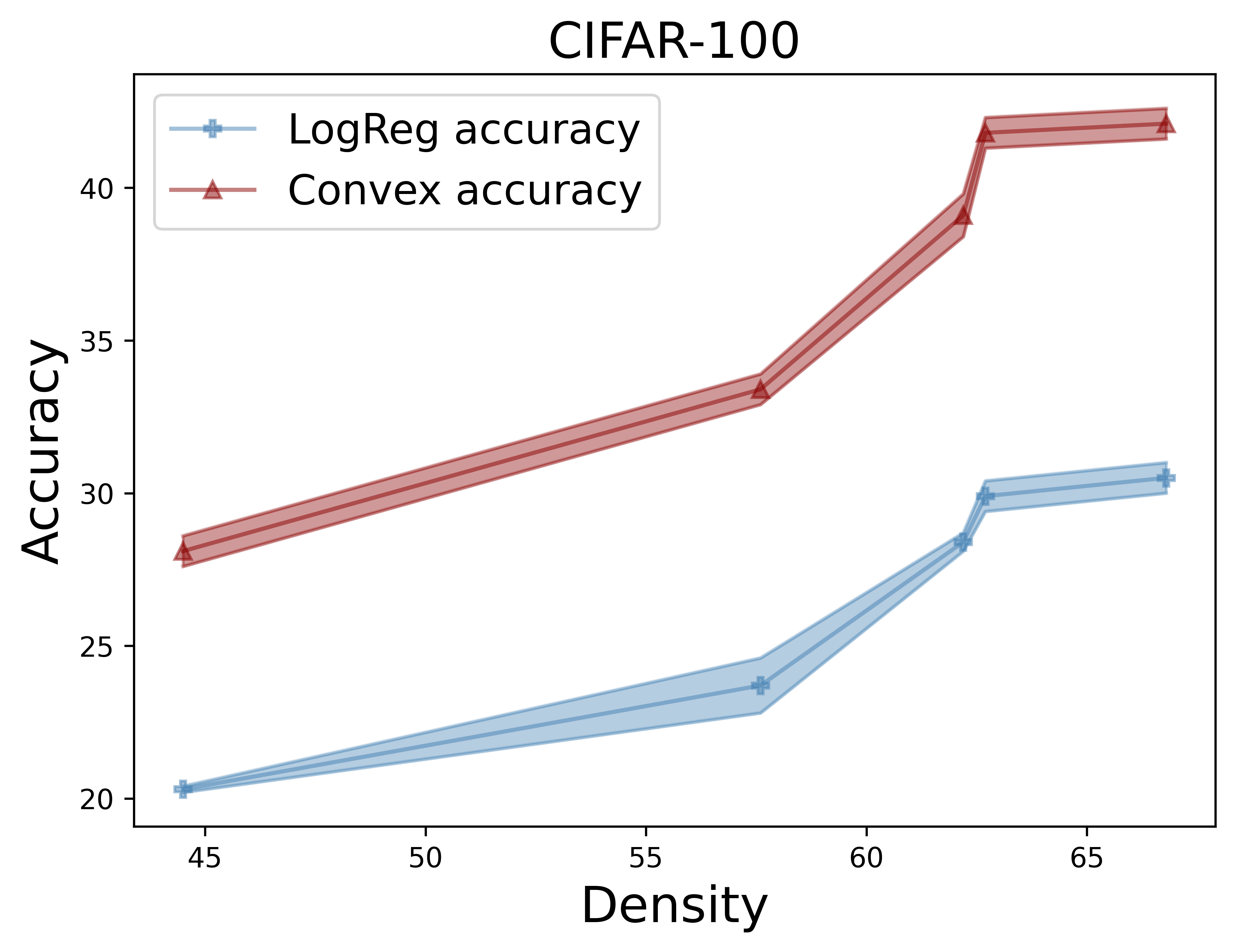}
    }\hfill
    \caption{Study of the correlation between GANs' performance and their latent space geometry. This is done by increasing the width of the generator ($w \in \{32,64,128,256,512\}$) in a fixed training setting on the CIFAR-10 and CIFAR-100 datasets. The results reveal a positive correlation between GANs' performance (measured by Precision and Density) and the linear separability and convexity of their latent space (measured by LogReg and Convex Accuracy). Confidence intervals are computed on 10 checkpoints of a training. \label{fig:correlation_study}}
\end{figure*}

\subsection{The latent geometry and GANs' performance}\label{sec:correlation_latent_geometry}
We investigate the relationship between the performance of GANs and their latent space geometry. To do so, we train many generators with different capacities (increasing width), and study how it impacts both the latent geometry and the performance. The results in Figure \ref{fig:correlation_study} reveal a strong positive correlation between the performance of GANs and the linearity/convexity of the latent space: the better the GANs perform, the more linearly separable and convex the latent space is. Indeed, the Pearson correlation between Precision and LogReg Accuracy is 0.98 on CIFAR-10, and 0.94 on CIFAR-100. Interestingly, overparametrization was known to help push-forward generative models in their optimization procedure \citep{balaji2020understanding} and in increasing their Lipschitz constant \citep{salmonacan}. We demonstrate here that it can help GANs  in reaching an optimal latent space structure, resulting in improved performance.

\subsection{Impact of the simplicial truncation method}\label{sec:impact_simplicial}
Finally, we aim to improve GANs performance by using our theoretical results (Theorem \ref{th:optimal_generators}). This is done by truncating the latent Gaussian distribution,   as discussed in Section \ref{sec:improvingGANs}, so that the generator structures its latent space with a simplicial cluster geometry. Note that the rejection threshold used at inference time can be higher than the one used at training time, since we have observed that higher rejection thresholds can help us increase both the precision and density of the models. The results in Table \ref{table:final_results} demonstrate that the use of this truncation method can improve the density and precision of GANs, without lowering the coverage nor the FID. This simplicial-based truncation has thus proved to be effective at removing off-manifold samples and can help improve push-forward generative models.

\section{Conclusion}
In conclusion, this paper takes a step towards a better understanding of push-forward generative models. When the latent space dimension is large enough, we prove the existence of an optimal latent space geometry, called `simplicial clusters'. Through experiments, we demonstrate that generative models with sufficient capacity tend to conform to this optimal geometry and also that enforcing this latent structure can improve GANs' performance. Our analysis has potential to drive further research on generative models with both theoretical and practical implications, such as developing new models that favor the emergence of such clusters in both latent and feature spaces. Similarly to what has been done in classification \citep{papyan2020prevalence}, studying thorougly the feature space of deep generative models is also an open question.


\begin{table}[h]
\small
\centering
\begin{tabular}{lccccccc} 
\multirow{2}{*}{Dataset/Model} & \multirow{2}{*}{\shortstack[c]{FID \\ $\downarrow$}} & \multirow{2}{*}{\shortstack[c]{Prec. \\  $\uparrow$}} & \multirow{2}{*}{\shortstack[c]{Rec. \\ $\uparrow$}} & \multirow{2}{*}{\shortstack[c]{Dens. \\ $\uparrow$}} & \multirow{2}{*}{\shortstack[c]{Cov.  \\ $\uparrow$}} \\ 
\\ \toprule
\textit{CIFAR-10} & & & & & \\ 

TransGAN & {\color{blue}8.9} & 72.8 & \textbf{62.6} & 79.3 & 79.3 \\ 
TransGAN + JBT & \textbf{8.8} & 73.3 & {\color{blue}61.2} & 85.7 & {\color{blue}81.1} \\ 
TransGAN + DeliG. & 9.8 & {\color{blue}74.6} & 58.6 & {\color{blue}93.2} & 80.0  \\
\textbf{TransGAN + simp.} & 9.2 & \textbf{74.9} & 59.2 & \textbf{96.4}  & \textbf{82.6} \\  \midrule
\textit{CIFAR-100} & & & & \\ 

TransGAN & 15.2  & 64.2 & \textbf{63.1} & 53.4 & 66.0 \\ 
TransGAN + JBT & \textbf{15.0} & {\color{blue}64.8} & {\color{blue}62.9} & {\color{blue}53.6} & {\color{blue}66.2} \\
TransGAN + DeliG. & 15.9 
& 63.5 & 62.2 
& 52.6 & 64.4 \\
\textbf{TransGAN + simp.} & {\color{blue}15.1} & \textbf{65.6} & 61.5 & \textbf{56.3} & \textbf{66.4} \\ \midrule  
\textit{STL-10 (32x32)} & & & & \\  

TransGAN & {\color{blue}10.5} & 75.7 & {\color{blue}60.1} & 87.5 & 83.0 \\
TransGAN + JBT & 11.0 
& \textbf{78.1}
& 57.6 
& \textbf{99.3}  
& \textbf{83.8}    \\
    TransGAN + DeliG. & {\color{blue}10.5} &  76.0 & \textbf{60.2} & 85.5 & 81.5  \\
\textbf{TransGAN + simp.} &\textbf{10.0} & {\color{blue}77.8} & {\color{blue}60.1} & {\color{blue}94.1} & {\color{blue}83.5}  \\  
\end{tabular}
\caption{Improving GANs with simplicial cluster latent space. JBT stands for the Jacobian-based truncation \citep{tanielian2020learning}; DeliG. for latent space with mixture of Gaussians  \citep{gurumurthy2017deligan}; 
simp. for our proposed truncation method with simplicial cluster. These results demonstrate that generators with a simplicial cluster latent space consistently outperform the baseline generator in Precision/Density, and most of the times outperforms other boosting methods (DeliGAN and JBT).
 \label{table:final_results}} 
\end{table} 

\paragraph{Limitations.} While our theoretical analysis demonstrates the existence of optimal generators, we were unable to prove their uniqueness. This limitation is associated with identifying partitions with the lowest $\varepsilon$-boundary measures in the Gaussian space, which is a challenging and unresolved problem in geometric measure theory.

\paragraph{Potential negative societal impacts. } This work may increase potential negative impacts of deep generative models, such as \textit{deepfakes}  \citep{fallis2020epistemic}. 

\clearpage
\newpage
\bibliography{main}

\appendix
\clearpage 
\newpage
\onecolumn

\appendix 

\section{Technical results: proofs} \label{appendix:proofs}


\subsection{Proof of Lemma \ref{lem:11}}
We want to show that generator $G \in \mathcal{G}_L^\mathcal{A}$ is such that $\alpha_G \leqslant 1 - \gamma (\partial^{\varepsilon_{\text{min}}} \mathcal{A})$, where
\begin{equation*}
    \partial^{\varepsilon_{\text{min}}} \mathcal{A} = \bigcup_{i=1}^m \big( \cup_{j \neq i} A_j\big)^{\varepsilon_{\text{min}}} \backslash \big(\cup_{j \neq i} A_j \big).
\end{equation*}

\textit{Proof by contradiction}. 

Assume a generator $G$ such that there exists $z \in \partial^{\varepsilon_{\text{min}}} \mathcal{A}$ and $i \in [1,m]$ such that $G(z) \in M_i$.
Since $G$ is associated with $\mathcal{A}$, we have using Definition \ref{def:association}, that there exists $z'$ and $j \in [1,m], j \neq i$ such that $\|z-z'\| < \varepsilon_{\text{min}}/2$ and 
$j = \underset{k \in [1,m]}{\argmin} \ \|G(z')-M_k\|$. Thus, we have:
\begin{align*}
 \|G(z)-G(z')\| 
    &\geqslant d(G(z'), M_i), \\
    &\geqslant d(M_i, M_i)/2, \\
    &\geqslant D_{\text{min}}/2. \\
    \text{And, } \frac{\|G(z)-G(z')\|}{\|z-z' \|} 
    &> D_{\text{min}} / \varepsilon_{\text{min}}, \\ &> L.
\end{align*}
This contradicts $G$ being in $\mathcal{G}_L^\mathcal{A}$. 

\subsection{Proof of Corollary \ref{cor:cor36}.}
Let $L, D$ be such that $L \geqslant D \sqrt{\log(m)}$. Let's prove that for any well-balanced generator $G \in \mathcal{G}_L$, we have:
    \begin{equation*}
        \alpha_G \leqslant 1 - \varepsilon_{\text{min}} \sqrt{\log m} \ e^{-3/2}. 
    \end{equation*}
    
    Using the method from \citet{schechtman2012approximate}, we have the measure of the border of cell $i$: 
\begin{align*}
    \gamma \Big( \big( \cup_{j \neq i} A_j\big)^\varepsilon \backslash \big(\cup_{j \neq i} A_j \big) \Big) 
    & \geqslant \frac{1}{\sqrt{2\pi}} \int_t^{t+\varepsilon} e^{-s^2/2} ds, \quad \text{where $t$ is such that $\frac{1}{\sqrt{2\pi}} \int_t^\infty e^{-s^2/2}ds = 1/m$}, \\
    & \geqslant \frac{\varepsilon}{\sqrt{2\pi}} e^{-(t+\varepsilon)^2/2}, \\
    & \geqslant \frac{\varepsilon \sqrt{\log m}}{m} e^{-\varepsilon t -\varepsilon^2/2} \quad (\text{using} \sqrt{\log m} \leq t \leq \sqrt{2 \log m}), \\
    & \geqslant \frac{\varepsilon \sqrt{\log m}}{m} e^{-\varepsilon \sqrt{\log m} -\varepsilon^2/2}.
\end{align*}

Thus:
\begin{align*}
    \gamma (\partial^{\varepsilon_{\text{min}}} \mathcal{A}) = \sum_{i=1}^m \gamma \Big( \big( \cup_{j \neq i} A_j\big)^\varepsilon \backslash \big(\cup_{j \neq i} A_j \big) \Big) 
    \geqslant \varepsilon_{\text{min}} \sqrt{\log m} \ e^{-\varepsilon_{\text{min}} \sqrt{\log m} -\varepsilon_{\text{min}}^2/2}. 
\end{align*}
Thus, we have
\begin{align*}
    \alpha_G 
    &\leqslant 1 - \gamma ( \partial^{\varepsilon_{\text{min}}} \mathcal{A}), \\
    &\leqslant 1 - \varepsilon_{\text{min}} \sqrt{\log m} \ e^{-\varepsilon_{\text{min}} \sqrt{\log m} -\varepsilon_{\text{min}}^2/2}.
\end{align*}

Moreover, using $\varepsilon_{\text{min}} = \frac{D}{L}$ and $L\geqslant D \sqrt{\log m}$, so we get $\varepsilon_{\text{min}} \sqrt{\log m} \leqslant 1$:
\begin{align*}
    \alpha_G \leqslant 1 - \varepsilon_{\text{min}} \sqrt{\log m} \ e^{-3/2}.
\end{align*}

\paragraph{Proof of Corollary \ref{cor:cor36} with $w_i$ (non-equal measure of modes).}

Let $L, D$ be such that $L \geqslant D \sqrt{\log(m)}$. Let's prove that for any well-balanced generator $G \in \mathcal{G}_L$, we have:
    \begin{equation*}
        \alpha_G \leqslant 1 - m w_\text{min} \varepsilon_{\text{min}} \sqrt{\log 1/w_\text{min}} \ e^{-3/2}.
    \end{equation*}
    
    Using the method from \citet{schechtman2012approximate}, we have the measure of the border of cell $i$: 
\begin{align*}
    \gamma \Big( \big( \cup_{j \neq i} A_j\big)^\varepsilon \backslash \big(\cup_{j \neq i} A_j \big) \Big) 
    & \geqslant \frac{1}{\sqrt{2\pi}} \int_t^{t+\varepsilon} e^{-s^2/2} ds, \quad \text{where $t$ is such that $\frac{1}{\sqrt{2\pi}} \int_t^\infty e^{-s^2/2}ds = w_\text{min}$}, \\
    & \geqslant \frac{\varepsilon}{\sqrt{2\pi}} e^{-(t+\varepsilon)^2/2}, \\
    & \geqslant w_\text{min}{\varepsilon \sqrt{\log 1/w_\text{min}}} e^{-\varepsilon t -\varepsilon^2/2} \quad (\text{using} \sqrt{\log 1/w_\text{min}} \leq t \leq \sqrt{2 \log 1/w_\text{min}}), \\
    & \geqslant  w_\text{min} {\varepsilon \sqrt{\log 1/w_\text{min}}} e^{-\varepsilon \sqrt{\log 1/w_\text{min}} -\varepsilon^2/2}.
\end{align*}

Thus:
\begin{align*}
    \gamma (\partial^{\varepsilon_{\text{min}}} \mathcal{A}) = \sum_{i=1}^m \gamma \Big( \big( \cup_{j \neq i} A_j\big)^\varepsilon \backslash \big(\cup_{j \neq i} A_j \big) \Big) 
    \geqslant m w_\text{min} \varepsilon_{\text{min}} \sqrt{\log 1/w_\text{min}} \ e^{-\varepsilon_{\text{min}} \sqrt{\log 1/w_\text{min}} -\varepsilon_{\text{min}}^2/2}. 
\end{align*}
Thus, we have
\begin{align*}
    \alpha_G 
    &\leqslant 1 - \gamma ( \partial^{\varepsilon_{\text{min}}} \mathcal{A}), \\
    &\leqslant 1 - m w_\text{min} \varepsilon_{\text{min}} \sqrt{\log 1/w_\text{min}} \ e^{-\varepsilon_{\text{min}} \sqrt{\log 1/w_\text{min}} -\varepsilon_{\text{min}}^2/2}.
\end{align*}

Moreover, using $\varepsilon_{\text{min}} = \frac{D}{L}$ and $L\geqslant D \sqrt{\log 1/w_\text{min}}$, so we get $\varepsilon_{\text{min}} \sqrt{\log 1/w_\text{min}} \leqslant 1$:
\begin{align*}
    \alpha_G \leqslant 1 - m \varepsilon_{\text{min}} w_\text{min} \sqrt{\log 1/w_\text{min}} \ e^{-3/2}.
\end{align*}

\subsection{Proof of Theorem \ref{th:optimal_generators}}
Let $\mu^\star$ be the target distribution. We know that $\mu_\star$ lays on m disconnected components contained in spheres $S_i, i \in [1,m]$.
We note $M_i, i \in [1,m]$ the centers, and $r_i$ the radius of each sphere. We also assume that the spheres verify Assumption \ref{ass:disconnect}. For each pair $(i,j) \in [1,m]^2$, we define $X_{ij} \in S_i$ and $X_{ji} \in S_j$
the points verifying
\begin{equation*}
    X_{ij} = \argmin_{x \in S_i} \ d(x, S_j) \quad \text{and} \quad
    X_{ji} = \argmin_{x \in S_j} \ d(x, S_i).
\end{equation*}

We consider the optimal partition $\mathcal{A}^\star$ in the Gaussian latent space. For each given latent point $z \in \mathds{R}^d$, we define:
\begin{equation*}
    N_z = \{ j \in [1,m], z \in A_j^\varepsilon \}.
\end{equation*}
We then distinguish two different cases:
\begin{enumerate}
    \item $|N_z| = 1$: the point $z$ belongs to the interior of a single cell, $z \in A_i^{-\varepsilon}$.
    \item $|N_z| \geqslant 2$: the point $z$ is in the neighborhood of at least two different cells.
\end{enumerate}

Interestingly, a point can only belong at most to the interior of one cell, but it can be at the intersection of several boundaries. We are now ready to define the optimal generator. 

\begin{figure}
\centering
\begin{tikzpicture}[scale=1.3]
\coordinate (O) at (0,0) ;
\coordinate (A) at (1,0.58) ;
\coordinate (B) at (-1,0.58) ;
\coordinate (C) at (0,-1.2) ;
\coordinate (F) at (-1,1.1) ;
\coordinate (I) at (0.1,0) ;
\coordinate (J) at (0.1,-1.5) ;
\draw (O) -- (A) ;
\draw (O) -- (B) ;
\draw (O) -- (C) ;
\draw [dashed,color=red] (0.05,0.086) -- (-0.95,0.666) ;
\draw [dashed,color=red] (-0.05,-0.086) -- (-1.05,0.494) ;
\draw [dashed,color=green] (-0.05,0.086) -- (0.95,0.666) ;
\draw [dashed,color=green] (0.05,-0.086) -- (1.05,0.494) ;
\draw [dashed,color=blue] (-0.1,0) -- (-0.1,-1.2) ;
\draw [dashed,color=blue] (0.1,0) -- (0.1,-1.2) ;
\draw  [fill=gray!20] (0,0) circle (0.1) ;

\draw (-0.5,-0.5) node[below]{$A_2$} ;
\draw (0.5,-0.5) node[below]{$A_3$} ;
\draw (0.,0.8) node[below]{$A_1$} ;

\draw (1.,-0.15) node[below]{$\partial^\varepsilon A_3$};

\draw (-0.2,-1.6) node[below]{Latent space} ;
\draw (4,-1.6) node[below]{Output space} ;

\coordinate (X1) at (4,0.7) ;
\coordinate (M2) at (3,-1) ;
\coordinate (M3) at (5,-1) ;

\fill[fill=gray!20] (X1)--(M2)--(M3);

\draw (M2) node[below]{$M_2$} ;
\draw (M3) node[below]{$M_3$} ;
\draw (X1) node[above]{$M_1$} ;

\draw [dashed,color=red] (X1) -- (M2) ;
\draw [dashed,color=green] (X1) -- (M3) ;
\draw [dashed,color=blue] (M2) -- (M3) ;

\draw [<-] (0.05,-0.3) -- (0.75,-0.3);
\draw [<-] (0.3,0.1) -- (0.75,-0.3);

\draw [->] (1.25,-0.3) --  (4.4,-0.1);
\draw [->] (1.25,-0.3) --  (4.0,-0.95);

\draw (X1) [fill=black] circle (0.03) ;
\draw (M2) [fill=black] circle (0.03) ;
\draw (M3) [fill=black] circle (0.03) ;
\end{tikzpicture}
\caption{ An optimal generator maps a 2D latent space to a 2D output space with three modes $(M_1,M_2,M_3)$. The latent space has an optimal ‘simplicial cluster' geometry. In the latent space, all the $\epsilon$-boundaries intersect each other in the gray circle, which is mapped in the output space in the convex hull of the three modes. \label{fig:optimal_G}}
\end{figure}
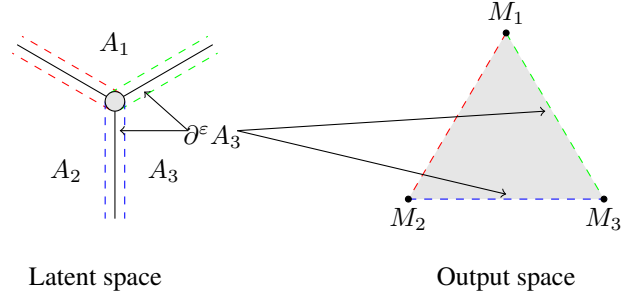

First, we set 
\begin{equation*}
    G(z) = X_{i,j} \ \text{for all} \ z \in \{z \in \mathds{R}^d, |N_z|=2, z \in \overline{A_i^{-\varepsilon}} \cap A_i^\varepsilon \cap A_j^\varepsilon \ \text{where} \ N_z = \{i,j\} \}.
\end{equation*}

Second, we define the generator in the interior of the cells, i.e. $N_z= \{i\}$. For each $z \in A_i^{-\varepsilon}$ and for a given unit vector $u \in \mathds{R}^d$, we assume that the generator is constant along the parametric line $z = k \times u, k \in \mathds{R}$. 

Finally, we define the generator when $z$ does not belong to the interior of any cell, i.e. $|N_z| \geqslant 2$: 
\begin{equation}\label{eq:Gstar2}
    G_\varepsilon^\star(z) = \sum_{i \in [1,m]} \sum_{j \neq i}
    w_{i,j}(z) \ X_{i,j} \ \mathds{1}_{j \in N_z} \ \mathds{1}_{i \in N_z}
    \quad \text{where} \quad 
    w_{i,j}(z) = \frac{d(z, (A_i^{\varepsilon})^\complement)}{\sum_{i \in [1,m]} \sum_{j \neq i} \ d(z, (A_j^{\varepsilon})^\complement) \ \mathds{1}_{j \in N_z} \ \mathds{1}_{i \in N_z}}
\end{equation}
where $d(z, A) = \min_{a \in A} \ \|z-a\|$.
An illustration of the optimal generator is given in Figure \ref{fig:optimal_G}. When z belongs to the intersection of two $\varepsilon$-boundaries, $G_\varepsilon(z)$ is a simple linear combination of 2 points. It is only when $|N_z| \geqslant 3$ that more complex samples are generated. A simple illustration of $G_\varepsilon^\star$ for $d=2$ and $m=3$ is given in Figure \ref{fig:optimal_G}. Interestingly, one can also show that the image of $G^\star_\varepsilon$ is equal to the convex hull of the Diracs $X_i, i \in [1,m]$. In particular, there exists a particularly interesting neighborhood $\nu$ of $0$ where $G^\star_\varepsilon(\nu)$ is equal to the whole convex hull of the points $X_i, i \in [1,m]$.

\paragraph{Proof that $G^\star_\varepsilon$ is well-balanced.}
    We recall that a generator is \textit{well-balanced} if we have $G \sharp \gamma (M_1) = \hdots = G \sharp \gamma (M_m)$. By construction \eqref{eq:Gstar2}, we have that for any $i \in [1,m]$ 
    \begin{align*}
        \|G^\star_\varepsilon(z) - X_i \| &= \| \sum_{k \neq i} w_k (X_k - X_i) \|, \\
        &= D \times (1-w_i).
    \end{align*}
    
    So, for any $z \in A_i$, we have that 
    \begin{equation*}
        i = \argmin_{j \in [1,m] \ w_j } = \argmin_{j \in [1,m] \ \|G(z)-X_j\|}     
    \end{equation*}
    Thus $G^\star_\varepsilon$ is associated with the optimal partition $\mathcal{A}^\star$, .
    
    Besides, for a given radius $r$ of the different modes, since everything is symmetrical, we have that $\gamma(\{z \in \mathds{R}^d, \|G(z)-X_1\| \leqslant r \} = \hdots = \gamma(\{z \in \mathds{R}^d, \|G(z)-X_m\| \leqslant r \}$. Thus, the generator is well-balanced. 
    
\paragraph{Showing that $G^\star_{\varepsilon^{\star}}$ is in $\mathcal{G}_L$.}
    
    

    It is clear that when $|N_z|=1$, we have that $G^\star_\varepsilon(z)$ is a $L$-Lipshitz continuous function. 
    
    Now, assume that $|N_z| \geqslant 2$. Consider $z, z'$ such that $N_z = N_z'$. Let $\alpha = (\alpha_1,\hdots, \alpha_m)$ and $\beta = (\beta_1, \hdots, \beta_m)$ be two vectors, both in $\mathds{R}^m$, such that for all $i \in [1, m]$:
    \begin{equation*}
        \alpha_i = \frac{d(z, (A_i^{\varepsilon})^\complement)}{\sum_{j \in \mathcal{A}_z} d(z, (A_j^{\varepsilon})^\complement)} \quad \text{and} \quad \beta_i = \frac{d(z', (A_i^{\varepsilon})^\complement)}{\sum_{j \in \mathcal{A}_z} d(z', (A_j^{\varepsilon})^\complement)} 
    \end{equation*}
    
    We have that 
    \begin{align*}
        \|G(z) - G(z')\| &= \| (1-\sum_{i \neq 1} \alpha_i) X_1 - (1-\sum_{i \neq 1} \beta_i) X_1 + \sum_{i \neq 1} \alpha_i X_i - \sum_{i \neq 1} \beta_i X_i \| \\
        &= \| \sum_{i \neq 1} (\alpha_i - \beta_i) (X_1 - X_i) \| \\
        & \leqslant \underset{(i,j) \in [1,m]^2}{\max} \|X_i - X_j\|  \ \|\alpha - \beta \|, \\
        & \leqslant \underset{(i,j) \in [1,m]^2}{\max} \|X_i - X_j\|  \ \|h(z) - h(z') \|,
    \end{align*}
    where $h$ is the function from $\mathds{R}^d \to \mathds{R}^m$ defined as:
    \begin{equation*}
        h(z) = (\frac{d(z, (A_1^{\varepsilon})^\complement)}{\sum_{i \in \mathcal{A}_z} d(z, (A_i^{\varepsilon})^\complement)}, \hdots, \frac{d(z, (A_m^{\varepsilon})^\complement)}{\sum_{i \in \mathcal{A}_z} d(z, (A_i^{\varepsilon})^\complement)}).
    \end{equation*}
    
    We can write $h = f \circ g$ with $f$ the function defined from $\mathds{R}^d \to \mathds{R}^m$ by
    \begin{equation*}
        f(z) = \Big(d(z, (A_1^{\varepsilon})^\complement), \hdots, d(z, (A_k^{\varepsilon})^\complement) \Big),
    \end{equation*}
    and $g$ the function defined on $\mathds{R}^m \setminus \{0\}$ by 
    \begin{equation*}
        g(z) = \frac{z}{\|z\|_1}
    \end{equation*}
    
    We have that $f$ is a $\sqrt{m}$-Lipschitz functions (given that $z \mapsto d(z, (A_m^{\varepsilon})^\complement)$ is $1$-Lipschitz). Besides, we know that outside the ball $B_{\varepsilon/2}(0)$, the function $g$ is $(2/\varepsilon)$-Lipschitz. Since it is clear that for every point $z$ such that $|N_z| \geqslant 2$, we have that $|f(z)| \geqslant \varepsilon/2$. Finally, the function $h$ is $\frac{2\sqrt{m}}{\varepsilon}$-Lipschitz.
    Thus, we have that:
    \begin{equation*}
        \|G^\star_{\varepsilon}(z) - G^\star_{\varepsilon}(z')\| \leqslant \frac{2D \sqrt{m}}{\varepsilon} \|z - z'\|,
    \end{equation*}
    with $D = \max_{i,j} \|X_i - X_j\|, (i,j) \in [1,m]^2, i \neq j$. 
    
    Now, by noting $\varepsilon_{\text{max}} = \ \frac{D}{L}$, and considering $\varepsilon^\star = 2\sqrt{m} \ \varepsilon_{\text{max}}$, we have:
    \begin{equation*}
        \|G^\star_{\varepsilon^\star}(z) - G^\star_{\varepsilon^\star}(z')\| \leqslant L \|z - z'\|.
    \end{equation*}
    
    Now, consider two latent vectors $z, z'$ in the same cell $\overline{A_i^{-\varepsilon}}$. There exists $i \in [1,m]$, and a pairs $(j,j') \in [1,m]^2$ (note that $j$ could be equal to $j'$) such that $G(z) = X_{i,j}$ and $G(z') =X_{i,j'}$. Using a similar reasoning as before, we can show that:
    \begin{equation*}
        \|G^\star_{\varepsilon^\star}(z) - G^\star_{\varepsilon^\star}(z')\| \leqslant L \|z - z'\|,
    \end{equation*}
    with $D = 2 \max_{i \in [1,m]} \ r_i$.

    We can now conclude on the Lipschitzness of $G^\star$ on $\mathds{R}^d$.

\paragraph{Proving that:    
for $m \leqslant d+1$, for any $\delta >0$, if $L$ is large enough, then, for any well-balanced $G \in \mathcal{G}_{L}$, we have $\alpha_{G^\star_{\varepsilon_{\text{max}}}} \geqslant \alpha_G - \delta $. }

Let $G$ be a well-balanced generator and $\mathcal{A}$ the partition associated with $G$. Let us first define the gaussian boundary measure $P_\gamma$ of a partition $\mathcal{A}$ of $\mathds{R}^d$. For partitions with smooth boundaries, it coincides with the $(d-1)$-dimensional gaussian measure of the boundary, defined as follows: 
    \begin{equation*}
        P_\gamma( \mathcal{A})  = {\lim \inf}_{\varepsilon \to 0} \frac{\gamma(\partial^\varepsilon \mathcal{A}) - \gamma(\mathcal{A})}{\sqrt{2/\pi}\varepsilon} 
    \end{equation*}
    Moreover, for sets with smooth boundaries, we have from \citet[Theorem 3.2.29]{federer2014geometric}: 
    \begin{equation*}
        {\lim \inf}_{\varepsilon \to 0} \frac{\gamma(\partial^\varepsilon \mathcal{A}) - \gamma(\mathcal{A})}{\sqrt{2/\pi}\varepsilon}  = {\lim }_{\varepsilon \to 0} \frac{\gamma(\partial^\varepsilon \mathcal{A}) - \gamma(\mathcal{A})}{\sqrt{2/\pi}\varepsilon}  
    \end{equation*}

Let us denote $\mathcal{A}^\star$, the optimal partition defined in \citet{milman2022gaussian}, based on simplicial  clusters. $A^\star$ is a standard partition where $\gamma (A^\star_1) = \hdots = \gamma (A^\star_m)$ for all i, and $\sum_i \gamma(A_i) = 1$. By the multi-bubble theorem \citep{milman2022gaussian}, simplicial clusters (such as $\mathcal{A}^\star$) are the unique minimizers of the gaussian isoperimetric problem, thus:
    \begin{align*}
        P_\gamma( \mathcal{A}^\star) &\leqslant P_\gamma( \mathcal{A}) \\
         {\lim}_{\varepsilon \to 0} \frac{\gamma(\partial^\varepsilon \mathcal{A}^\star)}{\varepsilon} &\leqslant  {\lim}_{\varepsilon \to 0} \frac{\gamma(\partial^\varepsilon \mathcal{A})}{\varepsilon} \\
         L_\mathcal{A} \leqslant L_{\mathcal{A}^\star}
    \end{align*}
    where $L_\mathcal{A} = {\lim}_{\varepsilon \to 0} \frac{\gamma(\partial^\varepsilon \mathcal{A}^\star)}{\varepsilon}$ and $L_\mathcal{A^\star} = {\lim}_{\varepsilon \to 0} \frac{\gamma(\partial^\varepsilon \mathcal{A}^\star)}{\varepsilon} $. 

    Then, for any $\delta>0$, there exists $\varepsilon'>0$ such that, for any $\varepsilon < \varepsilon'$,
    \begin{equation*}
        |  \frac{\gamma(\partial^\varepsilon \mathcal{A}^\star)}{\varepsilon} - L_{\mathcal{A}^\star} | < \delta
        \quad \text{,} \quad 
        | \frac{\gamma(\partial^\varepsilon \mathcal{A})}{\varepsilon}  - L_{\mathcal{A}} | < \delta 
        \quad \text{and} \quad
           L_{\mathcal{A}^\star}  \leqslant L_{\mathcal{A}}
    \end{equation*}
    Thus, for any $\delta>0$, there exists $\varepsilon'>0$ such that, for any $\varepsilon < \varepsilon'$,
    \begin{equation}\label{eq:for_thib}
        \gamma(\partial^\varepsilon \mathcal{A}^\star) \leqslant \gamma(\partial^\varepsilon \mathcal{A}) + 2\delta \varepsilon
    \end{equation}
    
    Besides, we know that 
    \begin{equation*}
        \alpha_G \leqslant 1 - \gamma(\partial^{\varepsilon_{\text{min}}} \mathcal{A}) 
    \end{equation*}
    Consequently, we have that:
    \begin{align*}
        \alpha_G & \leqslant 1 - \gamma(\partial^{\varepsilon_{\text{min}}} \mathcal{A}) \\ 
        &\leqslant 1 - \gamma(\partial^{\varepsilon_{\text{min}}} \mathcal{A}^\star) + 2 \delta \varepsilon_{\text{min}} \quad \text{using \eqref{eq:for_thib}}.
    \end{align*}

    Now, by construction of $G^\star_{\varepsilon_{\text{max}}}$, we have that
    \begin{equation*}
        \alpha_{G^\star_{\varepsilon_{\text{max}}}} \geqslant 1 - \gamma(\partial^{\varepsilon_{\text{max}}} \mathcal{A}^\star).
    \end{equation*}

    Consequently,     
    \begin{align*}
        \alpha_G & \leqslant 1 - \gamma(\partial^{\varepsilon_{\text{min}}} \mathcal{A}^\star) + 2 \delta \varepsilon_{\text{max}} +  \gamma(\partial^{\varepsilon_{\text{max}}} \mathcal{A}^\star) - \gamma(\partial^{\varepsilon_{\text{max}}} \mathcal{A}^\star) \\
        &\leqslant \alpha_{G^\star_\varepsilon} + 2 \delta \varepsilon_{\text{max}} + \gamma(\partial^{\varepsilon_{\text{max}}} \mathcal{A}^\star) - \gamma(\partial^{\varepsilon_{\text{min}}} \mathcal{A}^\star) \\
        &\leqslant \alpha_{G^\star_\varepsilon} + 2 \delta \varepsilon_{\text{max}} + \gamma(\partial^{\varepsilon_{\text{max}}} \mathcal{A}^\star) - 2 L_{\mathcal{A}^\star} {\varepsilon_{\text{max}}} - \gamma(\partial^{\varepsilon_{\text{min}}} \mathcal{A}^\star) + 2 L_{\mathcal{A}^\star} {\varepsilon_{\text{min}}} + 2 L_{\mathcal{A}^\star} ({\varepsilon_{\text{max}}}-{\varepsilon_{\text{min}}}) \\
        &\leqslant \alpha_{G^\star_\varepsilon} + 4 \delta \varepsilon_{\text{max}} + 2 L_{\mathcal{A}^\star} \varepsilon_{\text{max}}, \\
        &\leqslant \alpha_{G^\star_\varepsilon} + \varepsilon_{\text{max}} (4 \delta + 2 L_{\mathcal{A}^\star}).
    \end{align*}
    We conclude by choosing $L$ big enough such that $\varepsilon_{\text{max}}$ is strictly smaller than $\frac{\delta}{4 \delta + 2 L_{\mathcal{A}^\star}}$.


\paragraph{Proving the lower-bound \ref{eq:lower_bound_GANs} of Theorem \ref{th:optimal_generators}.}

Let's consider $G_{\varepsilon^\star}$ defined using \eqref{eq:Gstar2} and $\varepsilon^\star = 2 \sqrt{m} \varepsilon_{\text{max}}$. The precision of $G^\star_{\varepsilon^\star}$ is thus such that:
    \begin{equation*}
        \alpha_{G^\star_{\varepsilon^\star}} \geqslant 1 - \gamma(\partial^{\varepsilon^\star} \mathcal{A}).
    \end{equation*}
    However, since $\partial^\varepsilon \mathcal{A} \subset \bigcup_{i=1}^n A_i^\varepsilon$, we have that for any $\varepsilon$:
    \begin{align*}
        \gamma(\partial^\varepsilon \mathcal{A}) &\leqslant \sum_{i=1}^n \gamma(A_i^\varepsilon).
    \end{align*}
    Using results from \citet[][Proposition 1]{schechtman2012approximate}, when $m\leq d$, there exists $C$ large enough, such that
    \begin{align*}
        \gamma(A_i^{\varepsilon^\star}) 
        &\leqslant \frac{\varepsilon^\star}{m} \big(\sqrt{\pi \log(Cm)}\big).
    \end{align*}
    Thus, we have
    \begin{equation*}
        \alpha_{G^\star_{\varepsilon^\star}} \geqslant 1 - \varepsilon^\star \sqrt{\pi \log(Cm)}, 
    \end{equation*}
    To have $\alpha_{G^\star_{\varepsilon_{\text{max}}}} \geqslant 0$, we must have $^\star \leqslant 1/\sqrt{\pi \log(Cm)}$.
    This is the case since we have 
    \begin{align*}
        ^\star = 2 D \sqrt{m}/L \quad \text{and} \quad L \geqslant D \sqrt{m} \sqrt{\pi \log(Cm)},
    \end{align*}
    where $D = \max_{i,j} \ \|X_i-X_j\|$.

\section{Experiments} \label{appendix:experiments}
\subsection{Implementation details}
\begin{table*}[h]
\small
\centering
\caption{GANs training details on MNIST}
\begin{tabular}{llllll}

    \toprule
    Operation & Kernel & Strides & Feature Maps & Activation \\
    \midrule
    Generator G(z) \\ 
    $z \sim \mathrm{N}(0,I)$ & & & $dim(z)$ & \\
    Fully Connected & & & $7\times7\times128$ &  \\
    Convolution & $3\times3$ & $1\times1$ & $7\times7\times64$ & LReLU \\
    Convolution & $3\times3$ & $1\times1$ & $7\times7\times64$ & LReLU \\
    Nearest Up Sample & & & $14\times14\times64$ & \\
    Convolution & $3\times3$ & $1\times1$ & $14\times14\times32$ & LReLU \\
    Convolution & $3\times3$ & $1\times1$ & $14\times14\times32$ & LReLU\\
    Nearest Up Sample & & & $14\times14\times32$ & \\
    Convolution & $3\times3$ & $1\times1$ & $28\times28\times16$ & LReLU \\
    Convolution & $3\times3$ & $1\times1$ & $28\times28\times1$ & Tanh \\
    \midrule
    D(x) & & & $28\times28\times1$ & & \\
    Convolution & $4\times4$ & $2\times2$ & $14\times14\times512$ & LReLU \\
    Convolution & $3\times3$ & $1\times1$ & $14\times14\times512$ & LReLU \\
    Convolution & $4\times4$ & $2\times2$ & $7\times7\times512$ & LReLU \\
    Convolution & $3\times3$ & $1\times1$ & $7\times7\times512$ & LReLU \\
    Fully Connected & & & $1$ & - \\
    \midrule
    Batch size & 256\\
    Leaky ReLU slope & 0.2\\
    Gradient Penalty weight & 10\\
    Learning Rate Discriminator & $1\times 10^{-4}$ \\
    Learning Rate Generator & $5\times 10^{-5}$ \\
    Disciminator steps & 2 \\
    Optimizer & Adam &  $\beta_1:0.5$ &  $\beta_2:0.5$\\
    \bottomrule
\end{tabular}
\label{appendix:mnist_architecture}
\end{table*}

\begin{table*}
\centering
\small
\caption{GANs training details on CIFAR datasets. BN stands for batch-normalization.}
\begin{tabular}{llllll}
    \toprule
    & & & & Conditional & \\
    Operation & Kernel & Strides & Feature Maps & BN \citep{chen2018self} & Activation \\
    \midrule
    Generator G(z)\\ 
    $z \sim \mathrm{N}(0,Id)$ & & & 128 & & \\
    Fully Connected & & & $4\times4\times128$ & - &  \\
    ResBlock & $[3\times3]\times 2$ & $1\times1$ & $4\times4\times128$ & Y & ReLU \\
    Nearest Up Sample & & & $8\times8\times128$ & - & \\
    ResBlock & $[3\times3]\times 2$ & $1\times1$ & $8\times8\times128$ & Y & ReLU \\
    Nearest Up Sample & & & $16\times16\times128$ & - & \\
    ResBlock & $[3\times3]\times 2$ & $1\times1$ & $16\times16\times128$ & Y & ReLU \\
    Nearest Up Sample & & & $32\times32\times128$ & - & \\
    Convolution & $3\times3$ & $1\times1$ & $32\times32\times3$ & - & Tanh \\
    \midrule
    Discriminator D(x) & & & $32\times32\times3$ & & \\
    ResBlock & $[3\times3]\times2$ & $1\times1$ & $32\times32\times256$ & - & ReLU \\
    AvgPool & $2\times2$ & $1\times1$ & $16\times16\times256$ & - & \\
    ResBlock & $[3\times3]\times2$ & $1\times1$ & $16\times16\times256$ & - & ReLU \\
    AvgPool & $2\times2$ & $1\times1$ & $8\times8\times256$ & - & \\
    ResBlock & $[3\times3]\times2$ & $1\times1$ & $8\times8\times256$ & - & ReLU \\
    ResBlock & $[3\times3]\times2$ & $1\times1$ & $8\times8\times256$ & - & ReLU \\
    Mean spatial pooling & - & - & $256$ & - & \\
    Fully Connected & & & $1$ & - & - \\
    \midrule
    Batch size & 256\\
    Gradient Penalty weight & 10\\
    Learning Rate Discriminator & $1\times 10^{-4}$ \\
    Learning Rate Generator & $5\times 10^{-5}$ \\
    Discriminator steps & 2 \\
    Optimizer & Adam & $\beta_1=0.$ & $\beta_2=0.999$\\
    \bottomrule
\end{tabular}
\label{appendix:cifar_architecture}
\end{table*} 

First, let us note that we share our code in Supplementary Material for reproducibility.

\paragraph{Training.} We use the Wasserstein loss with gradient-penalty on interpolations of fake and real data. At each iteration, the discriminator is trained 2 steps and the generator 1 step with Adam optimizer. The batch size is 256. The learning rate of the discriminator is two times larger \citep{heusel2017GANs}, \textit{i.e.} $5\times 10^{-5}$ for the generator and $1\times 10^{-4}$ for the discriminator. GANs  are trained for 80k steps on MNIST and for 100k steps on CIFAR datasets. Architectures of generator and discriminator are described in Table \ref{appendix:mnist_architecture} and Table \ref{appendix:cifar_architecture}. 

For TransGAN \citep{jiang2021transgan}, we follow the implementation from the authors  available at \href{https://github.com/VITA-Group/TransGAN}{https://github.com/VITA-Group/TransGAN}. TransGAN is trained with a WGAN-GP loss, 4 discriminator steps for 1 generator step, and Adam optimizer with a learning rate of $10^{-4}$. 

\paragraph{Evaluation.} For evaluation metrics, we follow the setting proposed by the authors. For FID \citep{heusel2017GANs}, we use 50k real images and 50k fake images. For precision, recall, density and coverage \citep{kynkaanniemi2019improved,naeem2020reliable}, we use 10k real images and 10k fake images with nearest-k$=5$. 

\paragraph{GPUs.} For all datasets, the training of GANs was run on NVIDIA Tesla V100 GPUs (16 GB). The training of ResNet GANs for 100k steps on CIFAR takes around 30 hours. For TransGAN models, the training is done for 250k steps on two NVIDIA Tesla V100 GPUs, which takes around $35\times 2=70$ GPU hours.

\clearpage
\newpage 

\subsection{Correlation between latent space geometry and GANs' performance (Details for Section \ref{sec:correlation_latent_geometry})}
We present the full results of this study in Table \ref{table:inception_overparametrization_study}.
\begin{table*}[h]
\small
\centering
\begin{tabular}{lcccccccc} 
\multirow{2}{*}{Dataset} & \multirow{2}{*}{Width}  &  \multirow{2}{*}{\shortstack[c]{LogReg \\ Acc. $\uparrow$}} & \multirow{2}{*}{\shortstack[c]{Convex \\ Acc. $\uparrow$}} & \multirow{2}{*}{\shortstack[c]{FID  $\downarrow$}} & \multirow{2}{*}{\shortstack[c]{Prec.  $\uparrow$}} & \multirow{2}{*}{\shortstack[c]{Rec.  $\uparrow$}} & \multirow{2}{*}{\shortstack[c]{Dens.  $\uparrow$}} & \multirow{2}{*}{\shortstack[c]{Cov.  $\uparrow$}} \\ \\\toprule
\multirow{5}{*}{{\shortstack[c]{CIFAR-10 \\ (Resnet)}}} & 32 & 53.4 $\pm$ 0.5 & 61.1 $\pm$ 0.3  & 28.3 $\pm$ 0.6 
& 63.2 $\pm$ 0.6 
& 58.6 $\pm$ 0.9 
& 66.3 $\pm$ 1.5 
& 61.3 $\pm$ 1.1  \\
 & 64 & 60.7 $\pm$ 0.5 & 72.1 $\pm$ 0.6  & 20.6 $\pm$ 0.3 
& 65.7 $\pm$ 0.5 
& 62.0 $\pm$ 0.6 
& 71.4 $\pm$ 1.7 
& 71.5 $\pm$ 1.0 \\
& 128 & 63.4 $\pm$ 0.4  & 73.1 $\pm$ 0.6 & 17.0 $\pm$ 0.3 
& 65.9 $\pm$ 0.4 
& 64.5 $\pm$ 0.9 
& 71.2 $\pm$ 1.5 
& 74.6 $\pm$ 0.9  \\ 
& 256 & {65.0} $\pm$ 0.4 & \textbf{75.4} $\pm$ 0.4 &  \textbf{16.1} $\pm$ 0.3 
& 66.4 $\pm$ 0.5 
& \textbf{66.2} $\pm$ 1.0 
& 72.3 $\pm$ 1.5 
& 75.6 $\pm$ 1.0 \\ 
& 512 & \textbf{65.3} $\pm$ 0.6 & 75.2 $\pm$ 0.7 &  \textbf{16.1} $\pm$ 0.3 
& \textbf{66.8} $\pm$ 1.0 
& 66.1 $\pm$ 1.3 
& \textbf{72.8} $\pm$ 2.9 
& \textbf{76.1} $\pm$ 1.4 \\ \midrule
\multirow{5}{*}{{\shortstack[c]{CIFAR-100 \\ (Resnet)}}} & 32 & 20.3 $\pm$ 0.1 & 28.1 $\pm$ 0.5&  28.3 $\pm$ 0.3 
& 53.4 $\pm$ 0.7 
& 63.5 $\pm$ 0.8 
& 44.5 $\pm$ 1.3 
& 56.1 $\pm$ 1.2  \\
 & 64 & 23.7 $\pm$ 0.9 &  33.4 $\pm$ 0.5 & 23.4 $\pm$ 0.3 
& 59.9 $\pm$ 0.5 
& 64.6 $\pm$ 0.7 
& 57.6 $\pm$ 1.7 
& 67.6 $\pm$ 0.5 \\
 & 128 & 28.4 $\pm$ 0.3 & 39.1 $\pm$ 0.7 &  21.1 $\pm$ 0.4 
& 61.6 $\pm$ 0.4 
& 63.8 $\pm$ 0.5 
& 62.2 $\pm$ 1.0 
& 70.2 $\pm$ 0.5 \\
& 256 & 29.9 $\pm$ 0.5 & 41.8 $\pm$ 0.6 &  21.0 $\pm$ 0.4 
& 62.3 $\pm$ 0.6 
& \textbf{65.6} $\pm$ 0.6 
& 62.7 $\pm$ 2.0 
& 70.1 $\pm$ 0.9 \\
& 512 & \textbf{30.5} $\pm$ 0.5 & \textbf{42.1} $\pm$ 0.5 & \textbf{19.7} $\pm$ 0.4 
& \textbf{64.3} $\pm$ 0.8 
& 64.8 $\pm$ 0.8 
& \textbf{66.8} $\pm$ 1.9 
& \textbf{72.2} $\pm$ 0.6 \\
\end{tabular}
\caption{Correlation between GANs' performance and their latent space geometry.  Increasing the capacity of GANs tend to structure their latent space in simplicial clusters (better LogReg accuracy) and improve their performance on precision, density and coverage. Confidence intervals are computed on several sets of generated/training points from a given generator. \label{table:inception_overparametrization_study}}
\end{table*}

\clearpage
\newpage

\subsection{Details on simplicial truncation method (Details for Section \ref{sec:impact_simplicial})}
We provide here more details about our truncation method. First, the rejection sampling in the latent space $\mathds{R}^d$ of GANs procedure is the following: 
\begin{itemize}
    \item Define hyper-parameters threshold $\tau$, number of clusters $N$, latent space dimension $d$.
    \item Initialize N equidistant points in $\{(u_0,\dots,u_N) \mid u_i \in \mathds{R}^d\}$. This can be done easily when $N\leq d$.
    \item When sampling latent vectors $z \in \mathds{R}^d$, compute a softmax over the negative distances between $z$ and $u_i$: $p_i(z) = \frac{e^{-d(z,u_i)}}{\sum_j e^{-d(z,u_j)}}$. Then, z is selected if $\underset{i}{\max} \big(p_i(z)\big) > \tau$. 
\end{itemize}

Second, we add a classification loss to encourage the generator to use this latent structure. This loss is motivated by the need to maximize mutual information between the latent cluster and the modes of the generator \cite{khayatkhoei2018disconnected}, and can be written as:
\begin{equation*}
     L_c = - \mathbb{E}_{z \sim \gamma}[\ln q_\phi \big(i(z)|G_\theta(z)\big)]
\end{equation*}
where $q_\phi$ is parametrized by a second classification head added to the discriminator; $i(z) = \underset{i}{\argmax} \big(p_i(z)\big)$ is the index of the latent cluster of the sample. This loss is added during training, at each step of generator's and discriminator's training, during the first 20 epochs. It is then dropped, since we noticed that it harms the GANs performance if it is kept until the end of the training. 

Training hyper-parameters: for $N=10$, we use a latent dimension of $d=64$ and training threshold of $\tau=0.135$; for $N=100$, we use $d=128$ and $\tau=0.08$. 

During inference, if the generator has properly learned to use the different clusters of the latent space, we observe that augmenting the threshold $\tau$ leads to an increased density and precision. 

\begin{figure}[h]
    \centering
    {{   
        \includegraphics[width=0.3\linewidth]{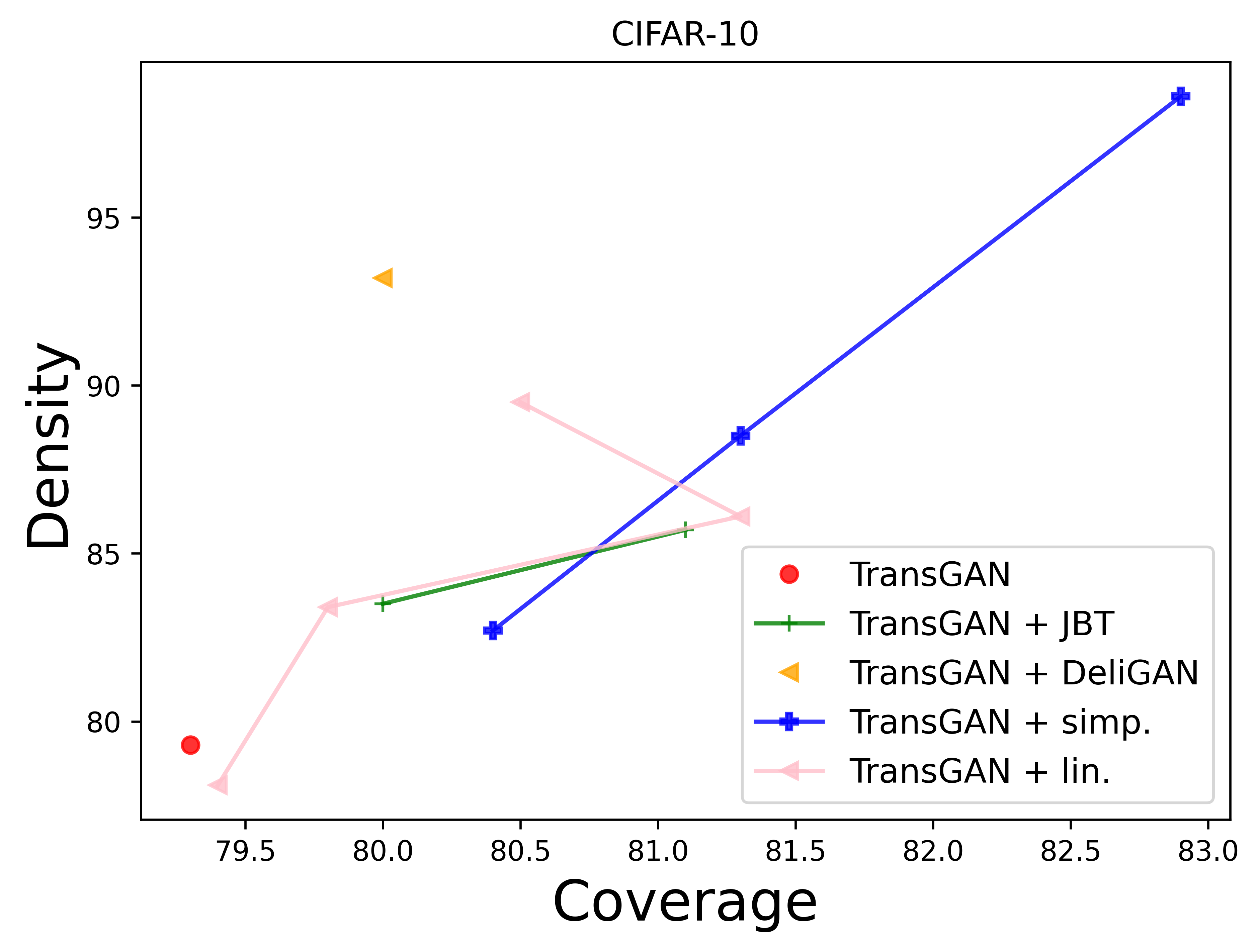}
    }}
    {{   
        \includegraphics[width=0.3\linewidth]{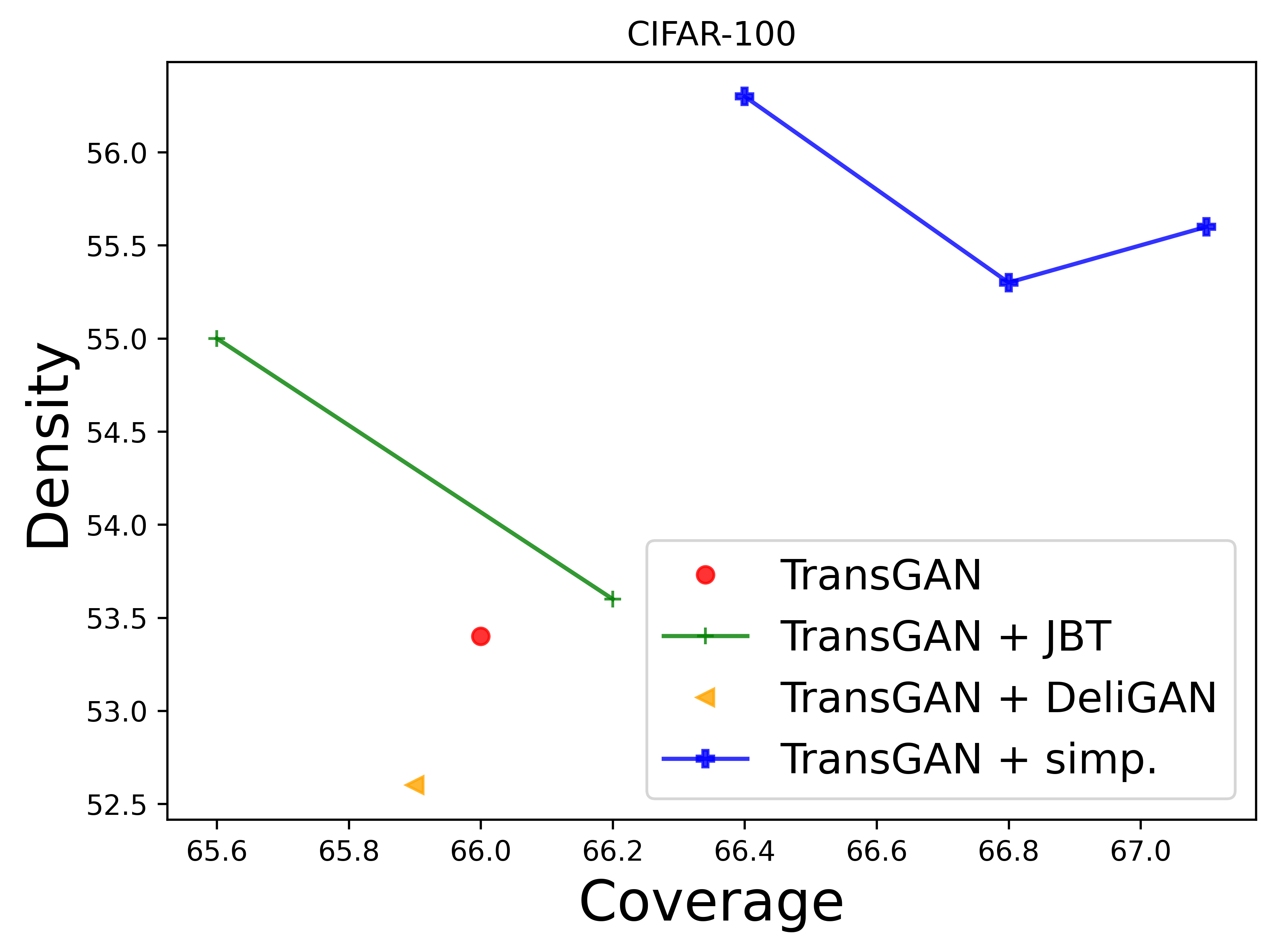}
    }}
    {{   
        \includegraphics[width=0.3\linewidth]{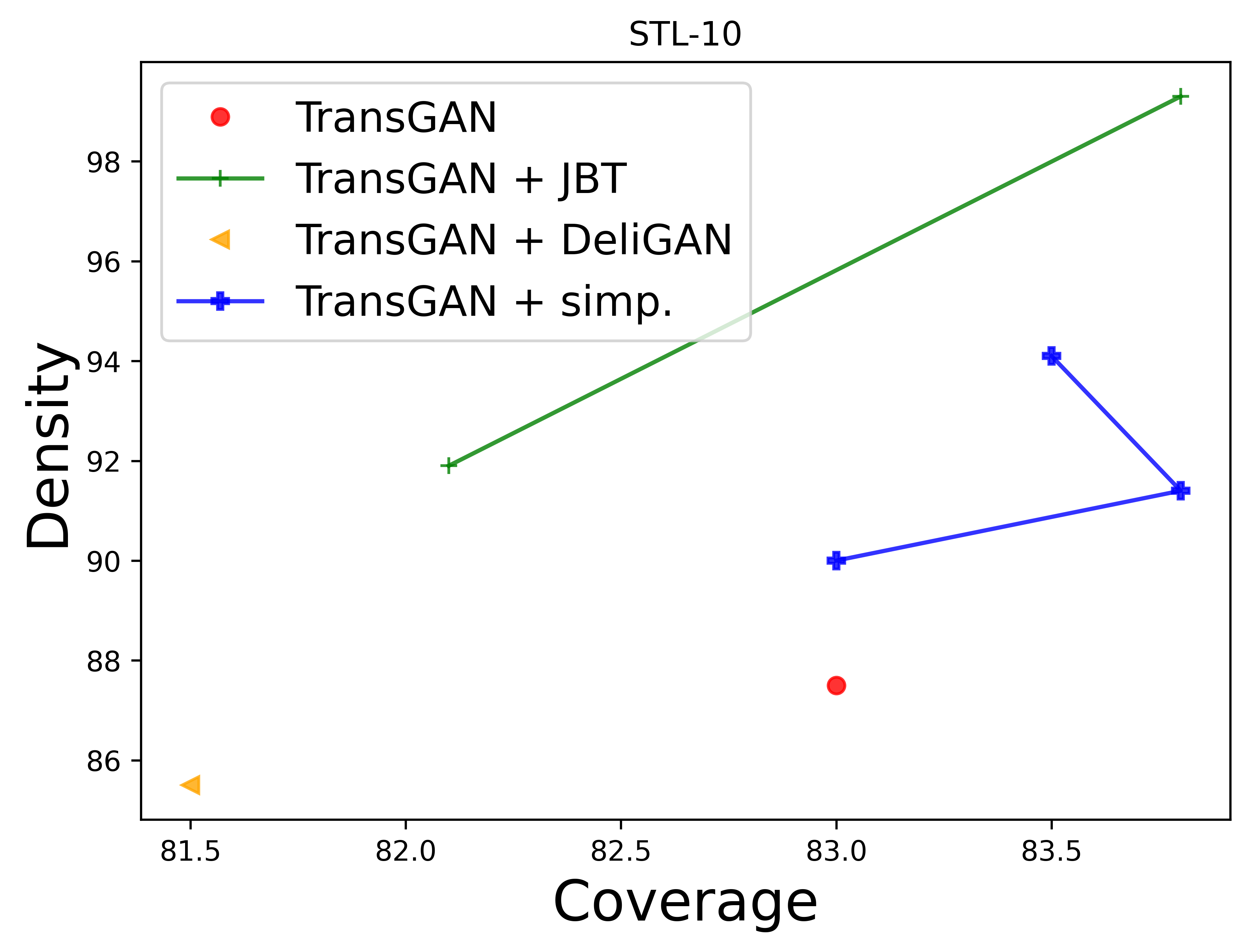}
    }}
    \caption{Density/Coverage curves comparing TransGAN and boosting methods for multi-modal datasets and different threshold ratios. Our simplicial truncation method (TransGAN + simp.) consistently outperforms the TransGAN and TransGAN + DeliGAN baselines. }
    \label{fig:simplicial_trunc_appendix}
\end{figure}

We present full results in Table \ref{tab:simplicial_trunc_appendix} and Figure \ref{fig:simplicial_trunc_appendix}.

\begin{table*}
\centering
\small
\begin{tabular}{lc|ccccc}
Dataset & Model   &  FID & Prec & Rec & Dens. & Cov.  \\ 
 \toprule
 \multirow{10}{*}{{\shortstack[c]{\textit{CIFAR-10} }}} &
 TransGAN & 8.9 $\pm$ 0.1  & 72.8 $\pm$ 0.8 & 62.6 $\pm$ 0.7 & 79.3 $\pm$ 0.9  & 79.3 $\pm$ 1.2   \\  \cmidrule(lr){2-7}
& TransGAN + 90\% JBT &8.7 $\pm$ 0.1 & 73.0 $\pm$ 0.6 &  61.9 $\pm$ 0.8 & 83.5 $\pm$ 1.7 & 80.0 $\pm$ 1.6 \\
& TransGAN + 80\% JBT & 8.8 $\pm$ 0.1 & 73.3 $\pm$ 0.8 & 61.2 $\pm$ 1.0
& 85.7 $\pm$ 2.8 & 81.1 $\pm$ 0.6 \\ \cmidrule(lr){2-7}
& TransGAN + DeliGAN N=10 & 9.8 $\pm$ 0.1 
 & 74.6 $\pm$ 0.8 & 58.6 $\pm$ 0.9 & 93.2 $\pm$ 2.8  & 80.0 $\pm$ 0.6 \\ \cmidrule(lr){2-7}
& TransGAN + lin. N=10 (0.23,0.23) & 9.2 $\pm$ 0.1 & 73.1 $\pm$ 1.1 & 61.9 $\pm$ 1.2 & 78.1 $\pm$ 2.7 & 79.4 $\pm$ 0.9 \\
& (0.23,0.29) &  9.2 $\pm$ 0.1 & 73.7 $\pm$ 0.8 & 61.5 $\pm$ 1.2 & 83.4 $\pm$ 3.3 & 79.8 $\pm$ 0.5 \\ 
 & (0.23,0.31) & 9.3 $\pm$ 0.1 & 74.0 $\pm$ 0.5 & 61.0 $\pm$ 0.7 & 86.1 $\pm$ 1.8 & 81.3 $\pm$ 0.7 \\
 & (0.23,0.4) & 9.8 $\pm$ 0.1 
& 75.1 $\pm$ 0.8 
& 59.8 $\pm$ 1.2 
& 89.5 $\pm$ 1.6 
& 80.5 $\pm$ 1.2 \\ \cmidrule(lr){2-7}
& TransGAN + simp. N=10, (0.135,0.135)  & 9.0 $\pm$ 0.1 & 72.9 $\pm$ 0.5  & 61.8 $\pm$ 0.9 & 82.7 $\pm$ 1.9  & 80.4 $\pm$ 0.7 \\ 
& (0.135,0.14) &  9.0 $\pm$ 0.1  & 74.2 $\pm$ 1.5  & 60.7 $\pm$ 1.0 & 88.5 $\pm$ 3.1  & 81.3 $\pm$ 1.4 \\
& (0.135,0.1445) &  9.3 $\pm$ 0.1& 75.3 $\pm$ 0.6 & 58.8 $\pm$ 0.8 & 98.6 $\pm$ 1.3 & 82.9 $\pm$ 0.5 \\


\midrule
\multirow{8}{*}{{\shortstack[c]{\textit{CIFAR-100} }}}   
& TransGAN & 15.2 $\pm$ 0.1 & 64.2 $\pm$ 0.5 & 63.1 $\pm$ 0.9 & 53.4 $\pm$ 1.3  & 66.0 $\pm$ 1.1\\ \cmidrule(lr){2-7}
& TransGAN + 90\% JBT &  15.1 $\pm$ 0.2 
& 64.8 $\pm$ 1.0 
& 62.9 $\pm$ 1.3 
& 53.6 $\pm$ 2.4 
& 66.2 $\pm$ 1.4 \\
& TransGAN + 80\% JBT &  14.8 $\pm$ 0.2 
& 65.4 $\pm$ 1.7 
& 61.7 $\pm$ 1.2 
& 55.0 $\pm$ 4.0 
& 65.6 $\pm$ 2.3 \\
\cmidrule(lr){2-7}
& TransGAN Deligan 10 &  15.9 $\pm$ 0.2
& 63.5 $\pm$ 0.8 & 62.2 $\pm$ 0.7 
& 52.6 $\pm$ 1.3 & 64.4 $\pm$ 0.6\\
& TransGAN DeliGAN 100 & 15.3 $\pm$ 0.1  
& 64.2 $\pm$ 0.5 
& 61.9 $\pm$ 0.9 
& 52.6 $\pm$ 0.6 
& 65.9 $\pm$ 0.8  \\ \cmidrule(lr){2-7}
& TransGAN + simp. N=10, (0.135,0.135) & 15.1 $\pm$ 0.1 
& 65.1 $\pm$ 0.6 
& 62.3 $\pm$ 0.5 
& 55.6 $\pm$ 0.6 
& 67.1 $\pm$ 0.5 \\
&  (0.135, 0.14) & 15.1 $\pm$ 0.1 
& 64.8 $\pm$ 0.2 
& 61.1 $\pm$ 0.5 
& 55.3 $\pm$ 1.3 
& 66.8 $\pm$ 1.1 \\
&  (0.135,0.1445)  & 15.1 $\pm$ 0.1 
& 65.6 $\pm$ 1.3 
& 61.5 $\pm$ 0.8 
& 56.3 $\pm$ 1.5 
& 66.4 $\pm$ 1.4 \\ \midrule 
\multirow{8}{*}{{\shortstack[c]{\textit{STL-10 (32x32)} }}} 
& TransGAN  & 10.5 $\pm$ 0.1 & 75.7 $\pm$ 0.6 & 60.1 $\pm$ 0.8 & 87.5 $\pm$ 1.9 & 83.0 $\pm$ 0.2 \\  \cmidrule(lr){2-7}
& TransGAN + 90\% JBT & 10.5 $\pm$ 0.1
& 76.9 $\pm$ 0.7 
& 58.8 $\pm$ 0.5 
& 91.9 $\pm$ 1.9 
& 82.1 $\pm$ 0.8 \\ 
& TransGAN + 80\% JBT & 11.0 $\pm$ 0.1 
& 78.1 $\pm$ 0.3 
& 57.6 $\pm$ 1.3 
& 99.3 $\pm$ 2.8 
& 83.8 $\pm$ 0.7 \\  \cmidrule(lr){2-7}
& TransGAN DeliGAN 10  & 12.1 $\pm$ 0.1 
& 74.2 $\pm$ 1.2 
& 60.2 $\pm$ 0.5 
& 81.5 $\pm$ 1.5 
& 79.6 $\pm$ 0.8 \\
& TransGAN DeliGAN 100  & 10.5 $\pm$ 0.2 
& 76.0 $\pm$ 0.5 
& 60.2 $\pm$ 1.6 
& 85.5 $\pm$ 2.8 
& 81.5 $\pm$ 1.4 \\ \cmidrule(lr){2-7}
& TransGAN + simp. N=100, (0.08,0.08)  & 10.1 $\pm$ 0.1 
& 76.5 $\pm$ 0.9 
& 60.2 $\pm$ 0.8 
& 90.0 $\pm$ 1.7 
& 83.0 $\pm$ 0.5 \\ 
& (0.08,0.15)  & 10.0 $\pm$ 0.1
& 76.9 $\pm$ 0.8 
& 59.9 $\pm$ 0.6 
& 91.4 $\pm$ 1.1 
& 83.8 $\pm$ 0.3 \\
& (0.08,0.20)  & 10.0 $\pm$ 0.1 
& 77.8 $\pm$ 0.6 
& 59.8 $\pm$ 0.8 
& 94.1 $\pm$ 0.9 
& 83.5 $\pm$ 0.8 \\

\end{tabular}
\caption{Density/Coverage curves comparing TransGAN and boosting methods for multi-modal datasets and different threshold ratios. Our simplicial truncation method (TransGAN + simp.) consistently outperforms the TransGAN and TransGAN + DeliGAN baselines.} \label{tab:simplicial_trunc_appendix}
\end{table*}

\clearpage 
\newpage

\subsection{Impact of the number of modes: a synthetic example (Details for Section \ref{sec:impact_latent_dim})}
\begin{figure}[h]
    \centering
    {   
        \includegraphics[width=0.4\linewidth]{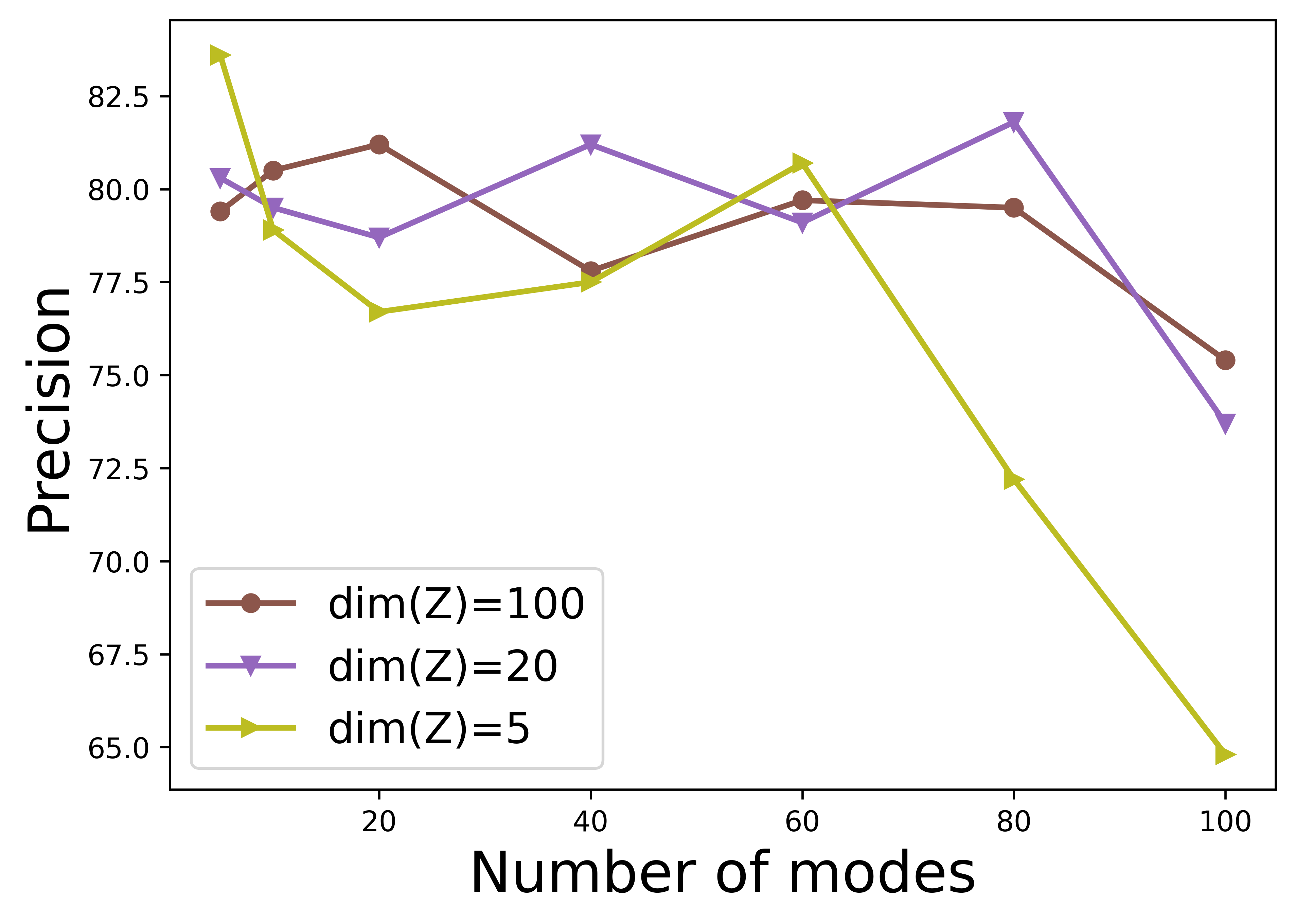}
    }
    {   
        \includegraphics[width=0.4\linewidth]{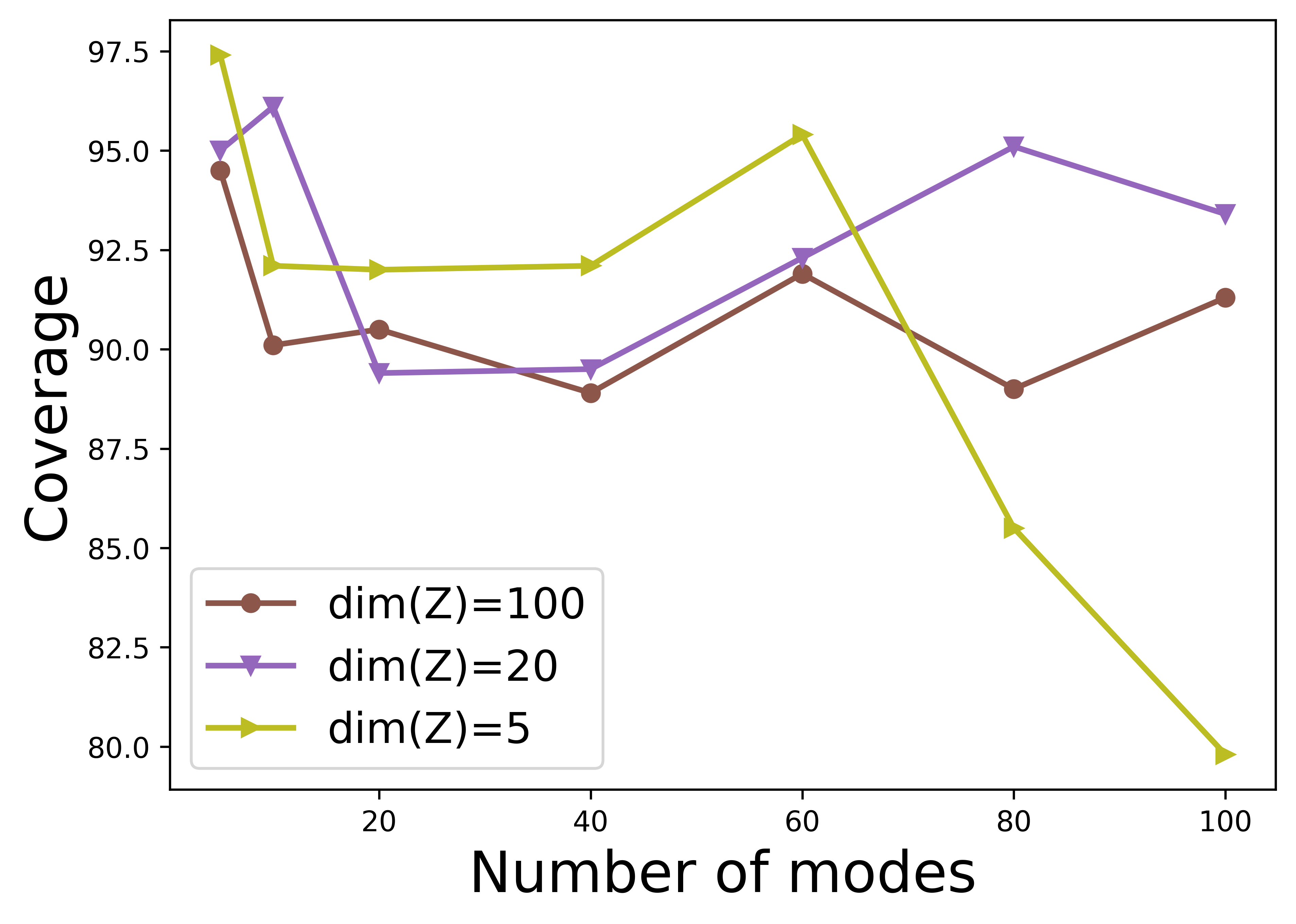}
    }
    \caption{Training on a mixture of Gausians in $\mathds{R}^{100}$ with varying number of modes and varying latent space dimension. The bigger the number of modes, the lower the precision. Increasing the latent space dimension helps up to a limit depending on the number of modes. 
    \label{fig:gaussian_mixture}}
\end{figure}

To illustrate our theoretical results, we propose to vary the number of modes of the data distribution. On real-world data, the number of modes is set but usually unknown, and removing/adding classes as a proxy for modes usually does not give insightful results since some classes can be much more complex than others. We thus use a synthetic setting, where we can easily control both the number of modes and their complexity. Figure \ref{fig:gaussian_mixture} stresses that as the number of modes increase, the precision decrease. Interestingly, using large latent space dimension can relieve the problem, even if the latent space dimension is clearly below that of the target. Recall the two problems that arise when training GANs: i) \textit{dimensional misspecification} where the true and modeled distributions do not have density functions w.r.t. the same base measure, and ii) \textit{density misspecification}, where GANs try to fit a disconnected manifold with a unimodal disitribution. From the results we conclude that:

\begin{itemize}
    \item With very low latent space dimensions, both problems i) and ii) have to be addressed and this leads to poor precision as the number of modes increases. 
    \item With larger latent space dimensions, the problem i) is less of a burden even when there is a clear dimensional misspecification and thus the GANs' performance is more tied to problem ii).
\end{itemize}

\end{document}